\documentclass[journal]{IEEEtran}

\usepackage{amssymb}
\usepackage{amsmath}
\usepackage{graphicx}
\usepackage{bbm}
\usepackage{booktabs}
\usepackage{multirow}
\usepackage{multicol}
\usepackage[flushleft]{threeparttable}
\usepackage{caption}
\usepackage[justification=centering]{subfig}
\usepackage{stackengine}
\usepackage{makecell}
\usepackage{balance}
\usepackage{fancyhdr}
\usepackage[square,sort,comma,numbers]{natbib}

\begin{document}


\title{Mutual Information-based Disentangled Neural Networks for Classifying Unseen Categories in Different Domains: Application to Fetal Ultrasound Imaging}

\author{Qingjie Meng,
        Jacqueline Matthew,
        Veronika A. Zimmer,
        Alberto Gomez,
        David F.A. Lloyd,  \\ 
        Daniel Rueckert,~\IEEEmembership{Fellow, ~IEEE},
        and Bernhard Kainz,~\IEEEmembership{Senior member, ~IEEE}
\thanks{This work is supported by the Wellcome Trust IEH [102431], Intel and Nvidia. We thank the ultrasound specialists Emily Skelton, Tom Day and David F.A. Lloyd.}
\thanks{Q. Meng, D. Rueckert and B. Kainz are with Imperial College London, SW7 2AZ, UK, (e-mail: q.meng16$\vert$bkainz$\vert$drueckert@imperial.ac.uk).}
\thanks{D. Rueckert is also with Faculty of Informatics and Medicine, Technical University Munich, Germany.}
\thanks{J. Matthew, V. Zimmer, A. Gomez, D. Lloyd are with King's College London, WC2R 2LS, UK. (e-mail: jacqueline.matthew$\vert$veronika.zimmer$\vert$alberto.gomez$\vert$david.lloyd@kcl.ac.uk)}
}

\markboth{ACCEPTED BY IEEE TRANSACTIONS ON MEDICAL IMAGING}%
{Qingjie Meng \MakeLowercase{\textit{et al.}}: Mutual Information-based Disentangled Neural Networks}

\maketitle


\begin{abstract}
Deep neural networks exhibit limited generalizability across images with different entangled domain features and categorical features. 
Learning generalizable features that can form universal categorical decision boundaries across domains is an interesting and difficult challenge.
This problem occurs frequently in medical imaging applications when attempts are made to deploy and improve deep learning models across different image acquisition devices, across acquisition parameters or if some classes are unavailable in new training databases.
To address this problem, we propose Mutual Information-based Disentangled Neural Networks (MIDNet), which extract generalizable categorical features to transfer knowledge to unseen categories in a target domain.
The proposed MIDNet adopts a semi-supervised learning paradigm to alleviate the dependency on labeled data. This is important for real-world applications where data annotation is time-consuming, costly and requires training and expertise.
We extensively evaluate the proposed method on fetal ultrasound datasets for two different image classification tasks where domain features are respectively defined by shadow artifacts and image acquisition devices. Experimental results show that the proposed method outperforms the state-of-the-art on the classification of unseen categories in a target domain with sparsely labeled training data.
\end{abstract}


\IEEEpeerreviewmaketitle

\section{Introduction}
\label{intro}
\IEEEPARstart{D}{eploying} deep neural networks (DNNs) in real-world clinical scenarios is challenging due to the problem of \emph{domain shift}~\cite{joaquin2009}. Domain shift corresponds to the feature distribution difference between training data and test data, which leads to performance degradation of DNNs from training to testing. This problem is ubiquitous in many clinical applications such as image classification~\cite{Navarro2018,Chen2019,Tang2019} and image segmentation~\cite{Kamnitsas2017,Jiang2018,Dou2019,Chartsias2019,Dou2020}. For example, performance degradation can be observed when applying a model that has been trained on images from one particular imaging device to images from another device. Addressing the problem of domain shift can contribute to a wider and efficient utilization of DNNs for image analysis in various clinical applications.

Domain shift can be categorized into (a) covariate shift (different latent feature distributions), (b) prior probability shift (change of labels) and (c) concept shift (different relationship between latent features and the desired label)~\cite{Saito2018,Lee2019}.
Covariate shift is the key reason for the lack of generalizability of DNNs. In medical imaging, covariate shift can be the result of the use of different imaging modalities (\emph{e.g.}, magnetic resonance imaging and ultrasound), different image acquisition devices within the same modality or different combinations of specific image features (\emph{e.g.}, anatomical structures and artifacts). 

\begin{figure}[t]
 \centering
 \includegraphics[width=0.49\textwidth, trim=3.2cm 12cm 12cm 1.8cm, clip]{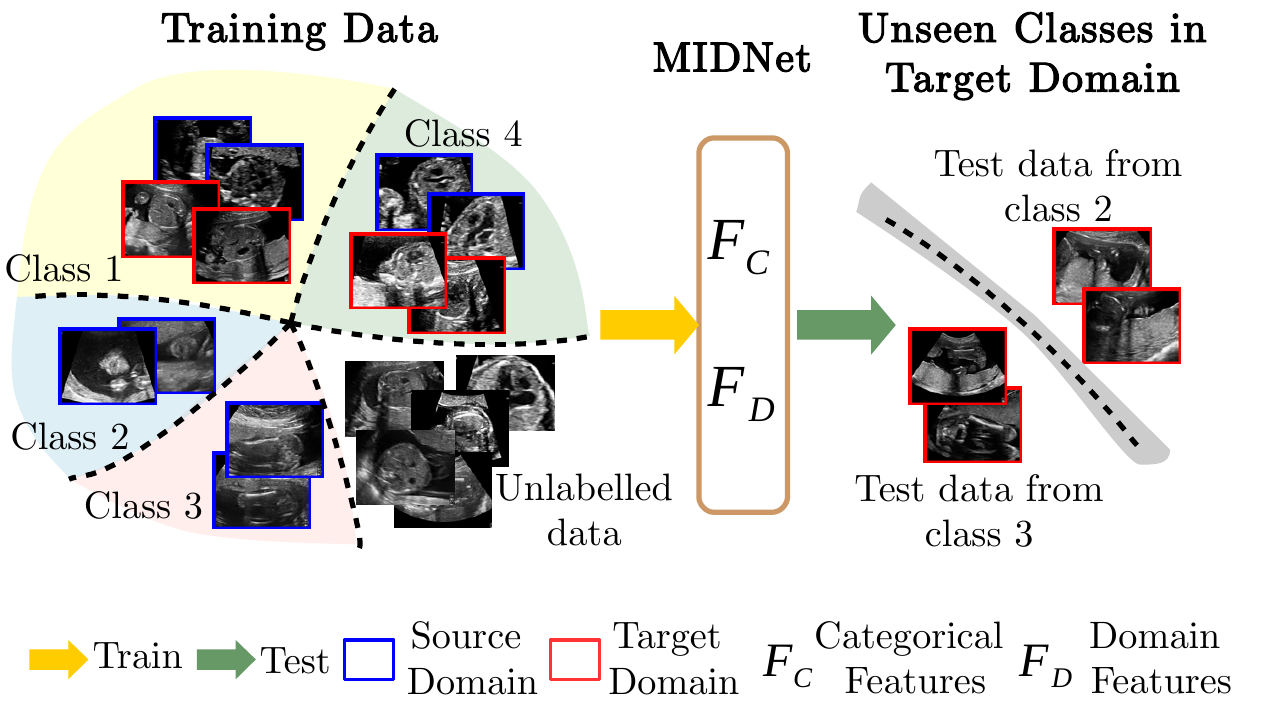}
 \caption{The proposed method (MIDNet) learns to extract generalized features ($\mathcal{F}_C$, $\mathcal{F}_D$) across domains from sparsely labeled training data. Generalized categorical features ($\mathcal{F}_C$) are able to correctly classify unseen categories in the target domain. This is of particular importance for real-world applications such as improving diagnostic classification in medical imaging, especially when some categories are unavailable in new datasets for training.}
 \label{flowchart}
\end{figure}

In contrast to the human visual system, DNNs exhibit weak generalizability when confronted with previously unseen entangled image features.
This is the problem which we address in this paper.
We postulate that DNNs should be able to learn generalizable features to transfer the knowledge from known entangled image features to new entangled image features. 
As outlined in Fig.~\ref{flowchart}, we want to improve the performance of DNNs on unseen categories in a target domain where all categories from a source domain and a subset of categories from a target domain are available for training.
This task can greatly contribute to diagnostic classification in medical imaging. For example, detecting a certain pathology which rarely occurs in a particular geographic region but might be common in other places. 

Fine-tuning DNNs on task-specific datasets is a possible solution but often infeasible due to the lack of sufficient annotated data in the target domain.
Domain adaptation algorithms have been widely studied to tackle the domain shift problem by extracting domain-invariant features (\emph{e.g.}, \cite{Kamnitsas2017,Jiang2018,Dou2019,Chartsias2019,Dou2020} in medical applications), aiming to transfer knowledge from a source domain to a target domain~\cite{Pan2010b}. 
Previous approaches can be categorized into three main groups: (1) \textit{Discrepancy measurement approaches} aim to align the feature distributions of source and target domain by measuring the discrepancy between representations, such as Maximum Mean Discrepancy (MMD)~\cite{Borgwardt2006mmd,Tzeng2014,Long2015} or correlation distance~\cite{Sun2016}; 
(2) \textit{Adversarial-based approaches} use DNNs to encourage the extracted features to be invariant for domain discrimination, instead of computing the discrepancy metric, and includes non-generative models~\cite{Ganin2016,Kamnitsas2017,Tzeng2017} and generative models~\cite{Liu2018, Bousmalis2017};  
(3) \textit{Reconstruction-based approaches} align the source and the target domain by image reconstruction which uses a cycle-consistency constraint to preserve domain-invariant features~\cite{Kim2017b,Hoffman2018}.


Existing domain adaptation methods can be practically prohibitive in real applications because a large amount of labeled data from source domains is needed. Although adversarial adaptation alternatives can perform well, optimizing adversarial objectives remains difficult and unstable in practice~\cite{Lezama2019}. Most importantly, previous methods make no explicit attempt to disentangle domain-invariant features from domain features, which results in the inability of dealing with previously unseen categories in the target domain.

In this paper, we propose mutual information-based disentangled networks (MIDNet) for representation disentanglement to address the problem outlined in Fig.~\ref{flowchart}.
The proposed approach extracts generalized categorical features by explicitly disentangling categorical features and domain features via mutual information minimization~\cite{Belghazi2018}. Note that the \textit{categorical features} in this paper refer to the features relevant to identities of classes or categories. We introduce the supervision from labeled images for an enhanced disentanglement via a feature clustering module, which estimates the similarity of categorical features from both domains. Image reconstruction is used to guarantee that the separated categorical and domain features are not random noise and are representative for the input images.
To further explore an improved categorical classification, we structure a categorical feature space by considering inter-class relationships. On top of the proposed MIDNet model, we incorporate distance metric learning to increase inter-class variance. 
The proposed method is a non-adversarial method which mitigates the difficulty and instability of adversarial model training.
Our method is a semi-supervised learning method, which only requires a small number of labeled samples during training while unlabeled data is integrated using a strategy similar to the MixMatch approach~\cite{Berthelot2019}.

\begin{figure}[tb]
 \centering
 \setcounter{subfigure}{0}
  \subfloat[Fetal US dataset]{
   \begin{tabular}{@{\hspace{-1\tabcolsep}}c@{\hspace{0.2\tabcolsep}}c@{\hspace{0.2\tabcolsep}}c@{\hspace{0.2\tabcolsep}}c@{\hspace{0.2\tabcolsep}}c@{\hspace{0.2\tabcolsep}}c}
   ~~~     &
   \raisebox{0.5\height}{\rotatebox[origin=c]{0}{\makecell{~\scalebox{0.8}{4CH}}}}    &
   \raisebox{0.5\height}{\rotatebox[origin=c]{0}{\makecell{~\scalebox{0.8}{Abdominal}}}}    &
   \raisebox{0.5\height}{\rotatebox[origin=c]{0}{\makecell{~\scalebox{0.8}{Femur}}}}    &
   \raisebox{0.5\height}{\rotatebox[origin=c]{0}{\makecell{~\scalebox{0.8}{Lips}}}}    &
   \raisebox{0.5\height}{\rotatebox[origin=c]{0}{\makecell{~\scalebox{0.8}{LVOT}}}}    \\
  \raisebox{1.5\height}{\rotatebox[origin=c]{90}{\makecell{~\scalebox{0.8}{SF}}}} &
  \includegraphics[height=1.2cm]{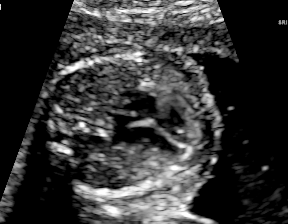} &
  \includegraphics[height=1.2cm]{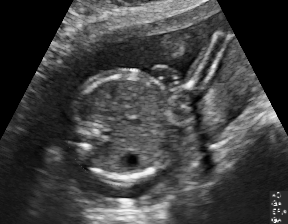} &
  \includegraphics[height=1.2cm]{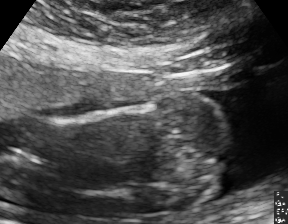} &
  \includegraphics[height=1.2cm]{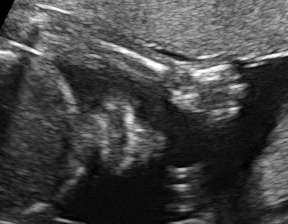} &
  \includegraphics[height=1.2cm]{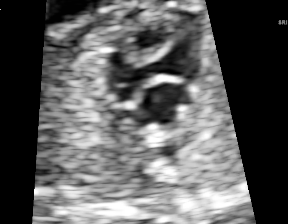} \\
  \raisebox{1.5\height}{\rotatebox[origin=c]{90}{\makecell{~\scalebox{0.8}{SC}}}} &
  \includegraphics[height=1.25cm, trim=2.4cm 1.4cm 2.4cm 1.4cm, clip]{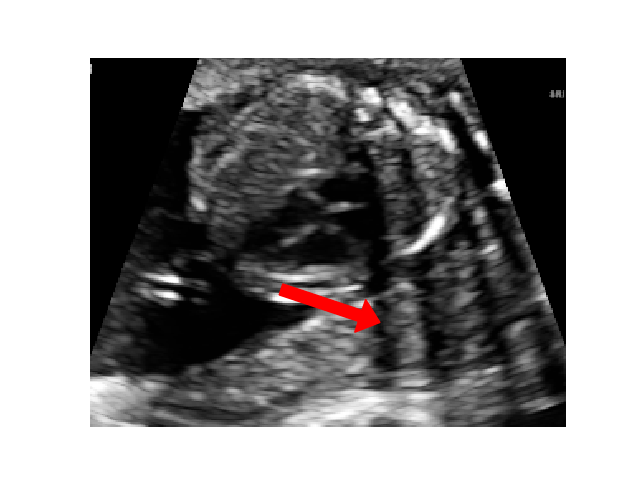} &
  \includegraphics[height=1.25cm, trim=2.4cm 1.4cm 2.4cm 1.4cm, clip]{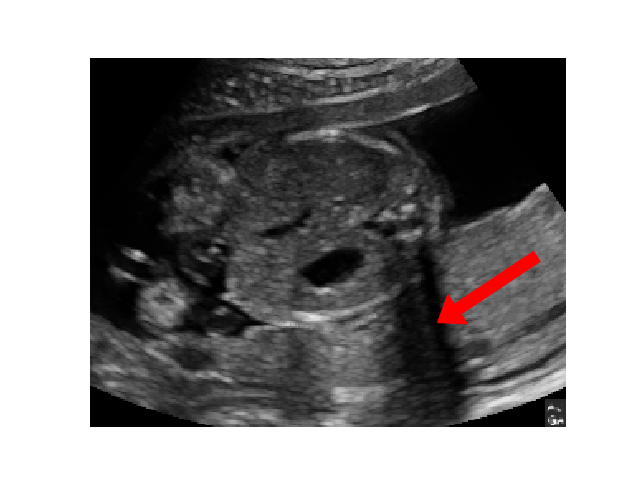} &
  \includegraphics[height=1.25cm, trim=2.4cm 1.4cm 2.4cm 1.4cm, clip]{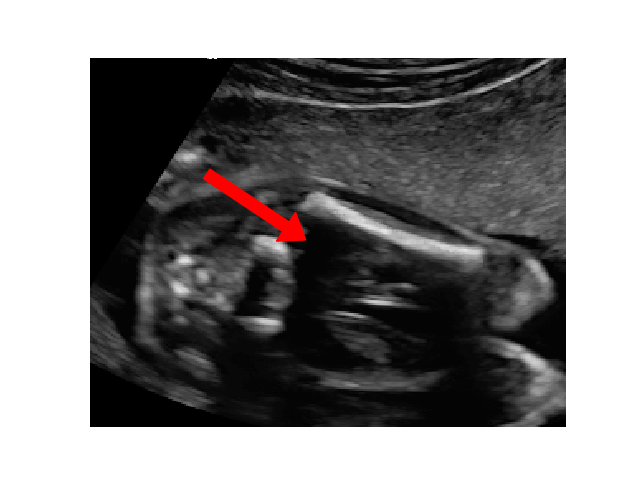} &
  \includegraphics[height=1.25cm, trim=2.4cm 1.4cm 2.4cm 1.4cm, clip]{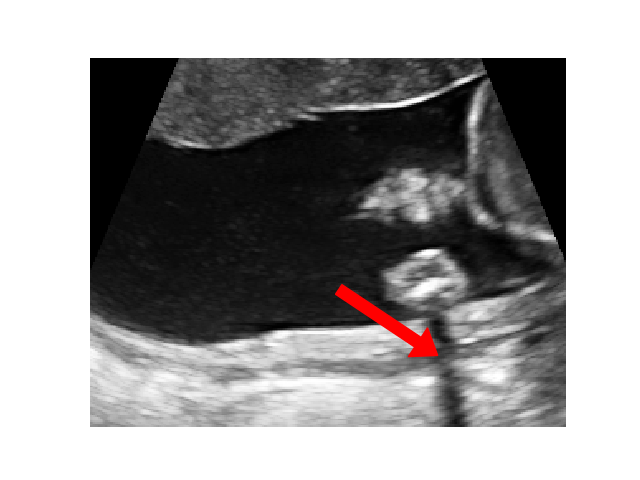} &
  \includegraphics[height=1.25cm, trim=2.4cm 1.4cm 2.4cm 1.4cm, clip]{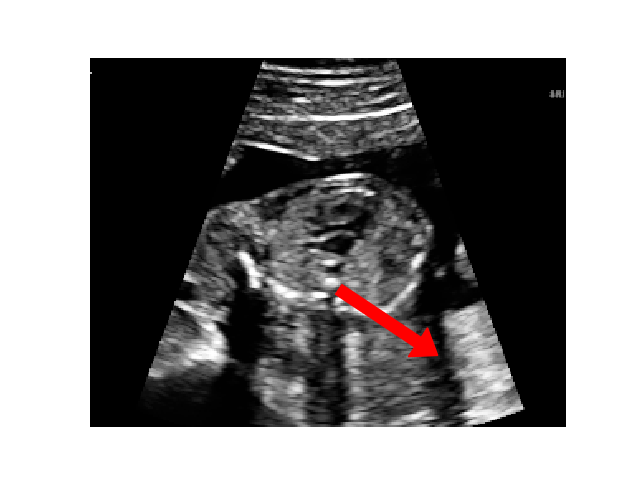} \\
  \raisebox{0.7\height}{\rotatebox[origin=c]{90}{\makecell{~\scalebox{0.8}{Histogram}}}} &
  \includegraphics[height=1.25cm, trim=2.2cm 1.2cm 4cm 2.5cm, clip]{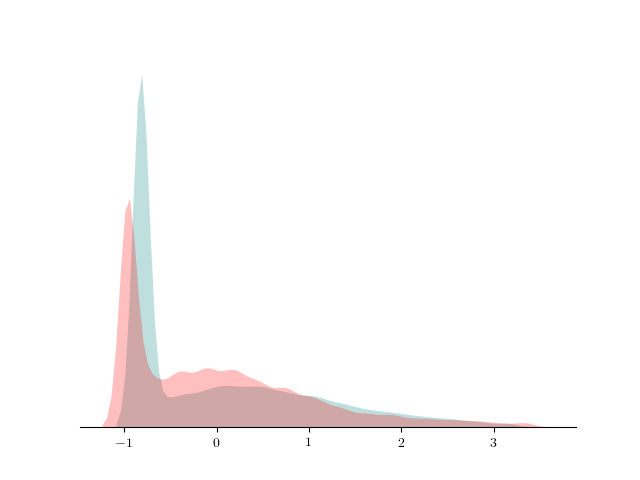} &
  \includegraphics[height=1.25cm, trim=2.2cm 1.2cm 4cm 2.5cm, clip]{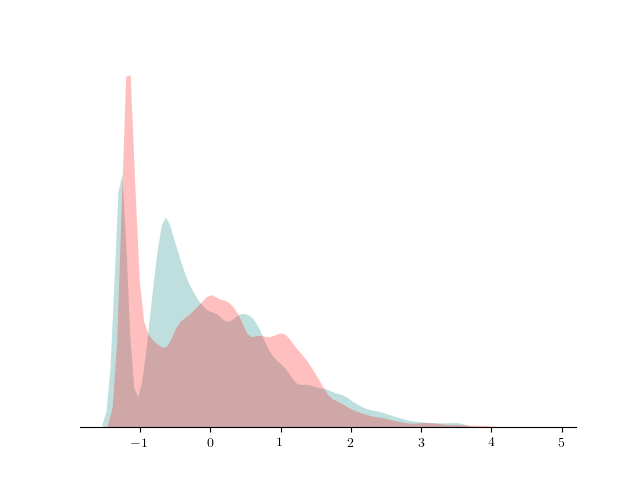} &
  \includegraphics[height=1.25cm, trim=2.2cm 1.2cm 4cm 2.5cm, clip]{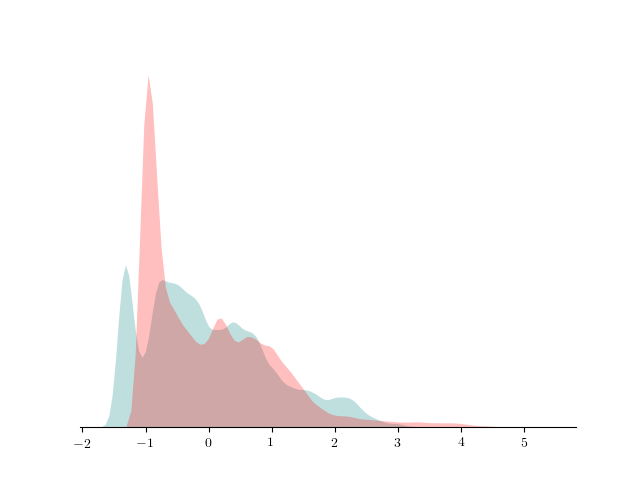} &
  \includegraphics[height=1.25cm, trim=2.2cm 1.2cm 4cm 2.5cm, clip]{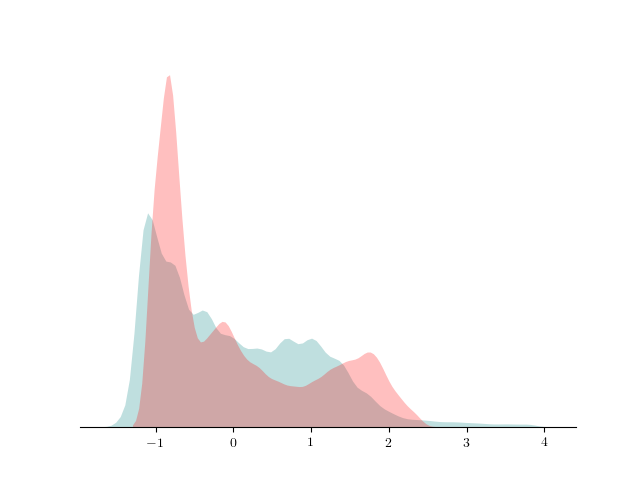} &
  \includegraphics[height=1.25cm, trim=2.2cm 1.2cm 4cm 2.5cm, clip]{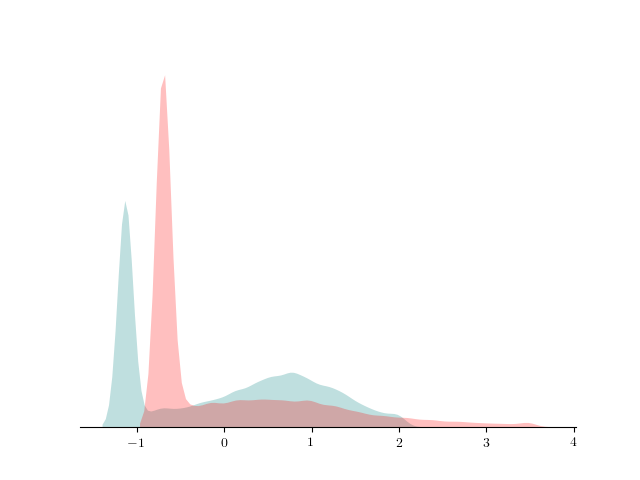}
  \end{tabular}
 }
  \\
 \setcounter{subfigure}{1}
  \subfloat[Cross-device dataset]{
   \begin{tabular}{c@{\hspace{0.6\tabcolsep}}c@{\hspace{0.6\tabcolsep}}c@{\hspace{0.6\tabcolsep}}c@{\hspace{0.6\tabcolsep}}c}
   ~~~     &
   \raisebox{0.5\height}{\rotatebox[origin=c]{0}{\makecell{~\scalebox{0.8}{Abdominal}}}}    &
   \raisebox{0.5\height}{\rotatebox[origin=c]{0}{\makecell{~\scalebox{0.8}{Brain}}}}    &
   \raisebox{0.5\height}{\rotatebox[origin=c]{0}{\makecell{~\scalebox{0.8}{Femur}}}}    &
   \raisebox{0.5\height}{\rotatebox[origin=c]{0}{\makecell{~\scalebox{0.8}{Lips}}}}    \\
  \raisebox{0.7\height}{\rotatebox[origin=c]{90}{\makecell{~\scalebox{0.8}{Device A}}}} &
  \includegraphics[height=1.25cm]{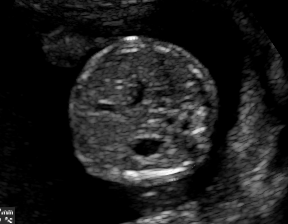}                              &
  \includegraphics[height=1.25cm]{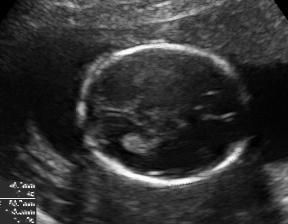} &
  \includegraphics[height=1.25cm]{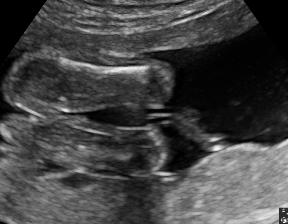}                              &
  \includegraphics[height=1.25cm]{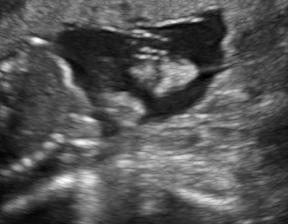}   \\
  \raisebox{0.7\height}{\rotatebox[origin=c]{90}{\makecell{~\scalebox{0.8}{Device B}}}} &
  \includegraphics[height=1.25cm]{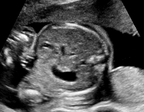} &
  \includegraphics[height=1.25cm]{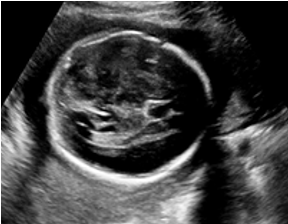}     &
  \includegraphics[height=1.25cm]{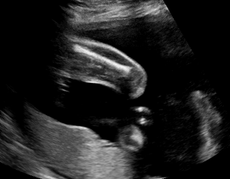} &
  \includegraphics[height=1.25cm]{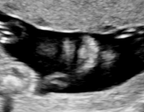} \\
  \raisebox{0.7\height}{\rotatebox[origin=c]{90}{\makecell{~\scalebox{0.8}{Histogram}}}} &
  \includegraphics[height=1.25cm, trim=2.2cm 1.2cm 4cm 2.5cm, clip]{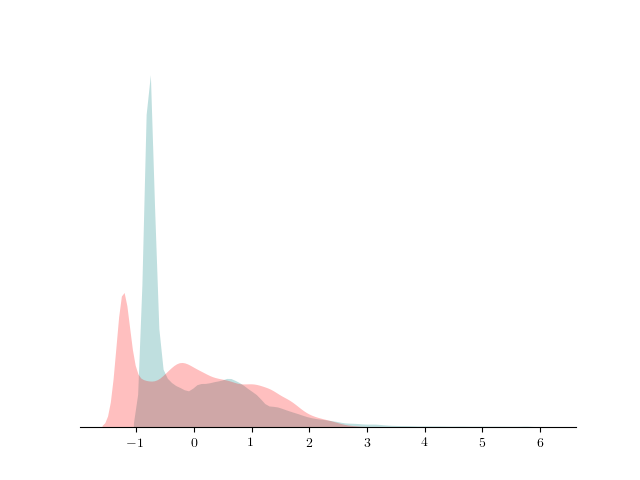} &
  \includegraphics[height=1.25cm, trim=2.2cm 1.2cm 4cm 2.5cm, clip]{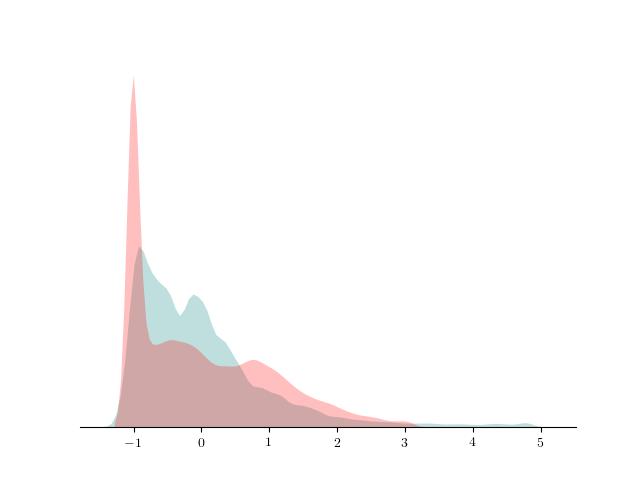} &
  \includegraphics[height=1.25cm, trim=2.2cm 1.2cm 4cm 2.5cm, clip]{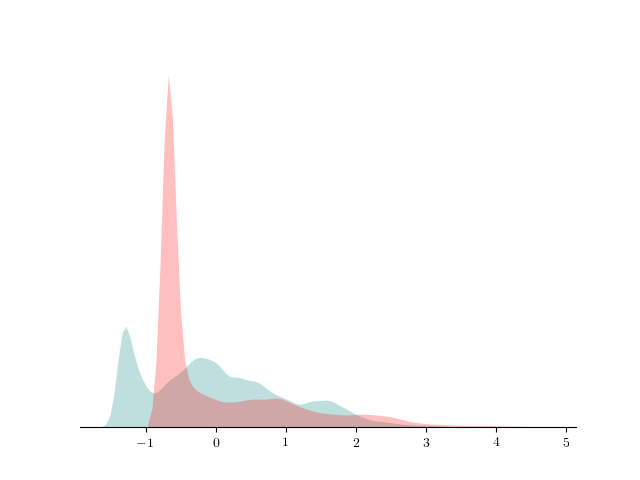} &
  \includegraphics[height=1.25cm, trim=2.2cm 1.2cm 4cm 2.5cm, clip]{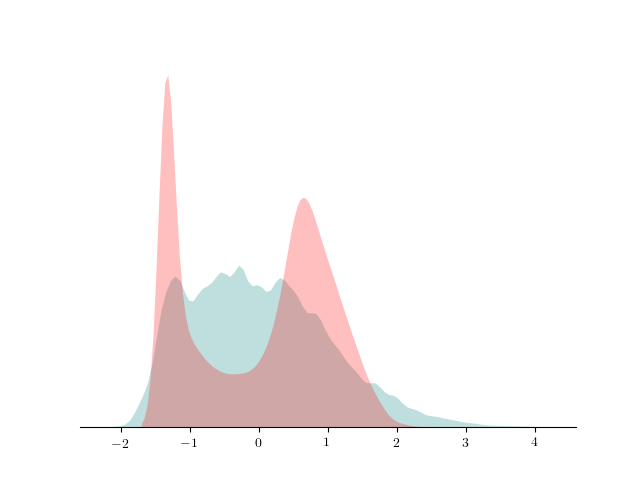}
  \end{tabular}
 }
  \caption{Examples of fetal US images and intensity histograms. (a) The fetal US dataset, including \textit{shadow-free} (SF) and \textit{shadow-containing} (SC) images of different anatomical structures. Red arrows show examples of acoustic shadow artifacts. (b) The fetal US dataset with images acquired by different devices, device A (GE Voluson E8) and device B ( Philips EPIQ V7 G).
  }
  \label{DataPresentation}
\end{figure}

We demonstrate the practical applicability of our method on a challenging medical application, the classification of standardized fetal ultrasound (US) views during prenatal screening. In many countries, US imaging is clinical routine for prenatal health care. 
The classification of standardized views is important for a consistent, cross-institutional  identification of possible abnormalities~\cite{screening2018}.
Early detection of pathological development can inform following treatment and delivery options~\cite{salomon2011, holland2015prenatal}. DNNs have shown promising performance to support this task~\cite{Baumgartner2017}.
However, ultrasound images are often hard to interpreted~\cite{Maraci2014}. Anatomical structures have inconsistent appearance~\cite{Baumgartner2017} and contain different orientations and shapes of anatomical landmarks~\cite{Ahmed2016}. Labeled training data is often insufficient as annotating medical images requires significant expertise and is prohibitively expensive in both time and labor. Manifestation of acoustic shadows~\cite{feldman2005,meng2018} as shown in Fig.~\ref{DataPresentation}(a) as well as different imaging devices as shown in Fig.~\ref{DataPresentation}(b) can lead to a domain shift problem for vanilla DNN classifiers. Exploring domain adaptation in fetal US enables DNN classifiers to be effectively utilized on a wider range, which supports identification of abnormalities from varying data sources. This can benefit prenatal healthcare. 

The main contributions of this paper are summarized as follows: 
\begin{itemize}
    \item We investigate a challenging domain adaptation problem for medical image classification: the translation of decision boundaries to a target domain, which lacks training samples for several categories. We propose end-to-end trainable Mutual Information-based Disentangled Networks (MIDNet) for learning generalized categorical features to classify unseen categories in the target domain.
    
    \item We develop MIDNet as a non-adversarial learning approach to show a more effective alternative to difficult and unstable adversarial training. Mutual information is utilized to separate categorical features from domain features, which is further supervised by labeled images via a feature clustering module. Image reconstruction is introduced to ensure the separated features are representative and meaningful for the input images.
    
    \item The proposed method extends the body of literature about semi-supervised domain adaptation (SSDA), which integrates unlabeled data from both source and target domain to alleviate the demand for annotated data, and thus MIDNet can be considered a new SSDA variant.
    
    \item We utilize our method for anatomical classification in fetal US, which, to the best of our knowledge, is the first exploration of transferring knowledge to unseen data in a practical application in medical imaging.
\end{itemize}

\section{Related work}
\subsubsection{Representation disentanglement} 
Disentangling representations aims at interpreting underlying interacted factors within data~\cite{Bengio2013,Chen2017} and enables the manipulation of relevant representations for specific tasks~\cite{Garcia2018,LiuA2018,Hadad2018}. Traditional models include techniques such as Independent Component Analysis (ICA)~\cite{Hyvarinen2000} and bilinear models~\cite{Tenenbaum2000} as well as learning-based models such as InfoGAN~\cite{Chen2016} and $\beta$-VAE~\cite{Higgins2017,Burgess2018}.
Recent work by Mathieu et al.~\cite{Mathieu2016} proposes a conditional generative model to disentangle latent representations into specified and unspecified factors of variation via adversarial training. 
For the same task, Hadad et al.~\cite{Hadad2018} proposes a simpler two-step adversarial training approach for more efficient learning of various unspecified features. Their method directly utilizes the encoded latent space for unspecified factors instead of assuming the underlying distribution. On top of adversarial training, Peng et al.~\cite{Peng2019} adds mutual information to disentangle more specific features, including class-irrelevant features, domain-invariant features and domain specific features.
To use disentangled representations for identifying images with unseen entangled features in real applications, Meng et al.~\cite{meng2019} proposed to disentangle category and domain-specific features using an adversarial regularization in a multi-task learning framework. 
In contrast to these previous works, we propose a non-adversarial method that evaluates mutual information between latent features to disentangle categorical features and domain features (Sec.~\ref{mutual}). Additionally, our method uses sparsely labeled data during training.    

\subsubsection{Semi-supervised learning (SSL)} The goal of SSL is to address the scarcity of labeled data by leveraging unlabeled data. Various approaches have been proposed for SSL~\cite{Chapelle2006,Lee2013,Laine2017,Miyato2018,Zhang2018,ZhangG2019}. Recently, Zhang et al.~\cite{Zhang2018} proposed MixUp as learning paradigm to train a model on convex combinations  of samples and their corresponding labels. This principle encourages the model to favor linear behavior between samples and alleviates problems arising from mislabelled examples. Extending this work, Berthelot et al.~\cite{Berthelot2019} introduced a SSL method, MixMatch, which estimates low-entropy labels for unlabeled samples and then applies MixUp to mixed labeled and unlabeled samples for training the model. In this paper, we utilize MixMatch to integrate unlabeled samples from both source and target domain during training (Sec.~\ref{mixmat}).

\subsubsection{Domain adaptation} 
Previous domain adaptation approaches consider three domain adaptation settings regarding to the number of domains, including one-to-one domain adaptation, multi-source domain adaptation and multi-target domain adaptation. One-to-one domain adaptation considers a single source domain and a single target domain. Unsupervised domain adaption~\cite{Ganin2016,Lee2019,mengMetFA2020} is a typical one-to-one domain adaptation, which requires plenty of labeled samples from the source domain during training and categories in both domains are the same. Multi-source domain adaptation learns universal knowledge from multiple source domains to a single target domain~\cite{peng2019moment}. Domain generalization~\cite{Dou2019,Li2019} is a special case of multi-source domain adaptation, which learns knowledge for an unseen target domain from many labeled samples of multiple source domains. In domain generalization, each category in the target domain has been seen in at least one source domain. Multi-target domain adaptation learns knowledge from a single source domain to multiple target domains. Domain agnostic learning~\cite{Peng2019} is a multi-target domain adaptation method, which also requires plenty of labeled samples from the source domain. In domain agnostic learning, all categories in target domains have been seen in the source domain.

\begin{figure}[pt]
 \centering
  \includegraphics[width=0.49\textwidth, trim=5.2cm 7.8cm 10.5cm 1cm, clip]{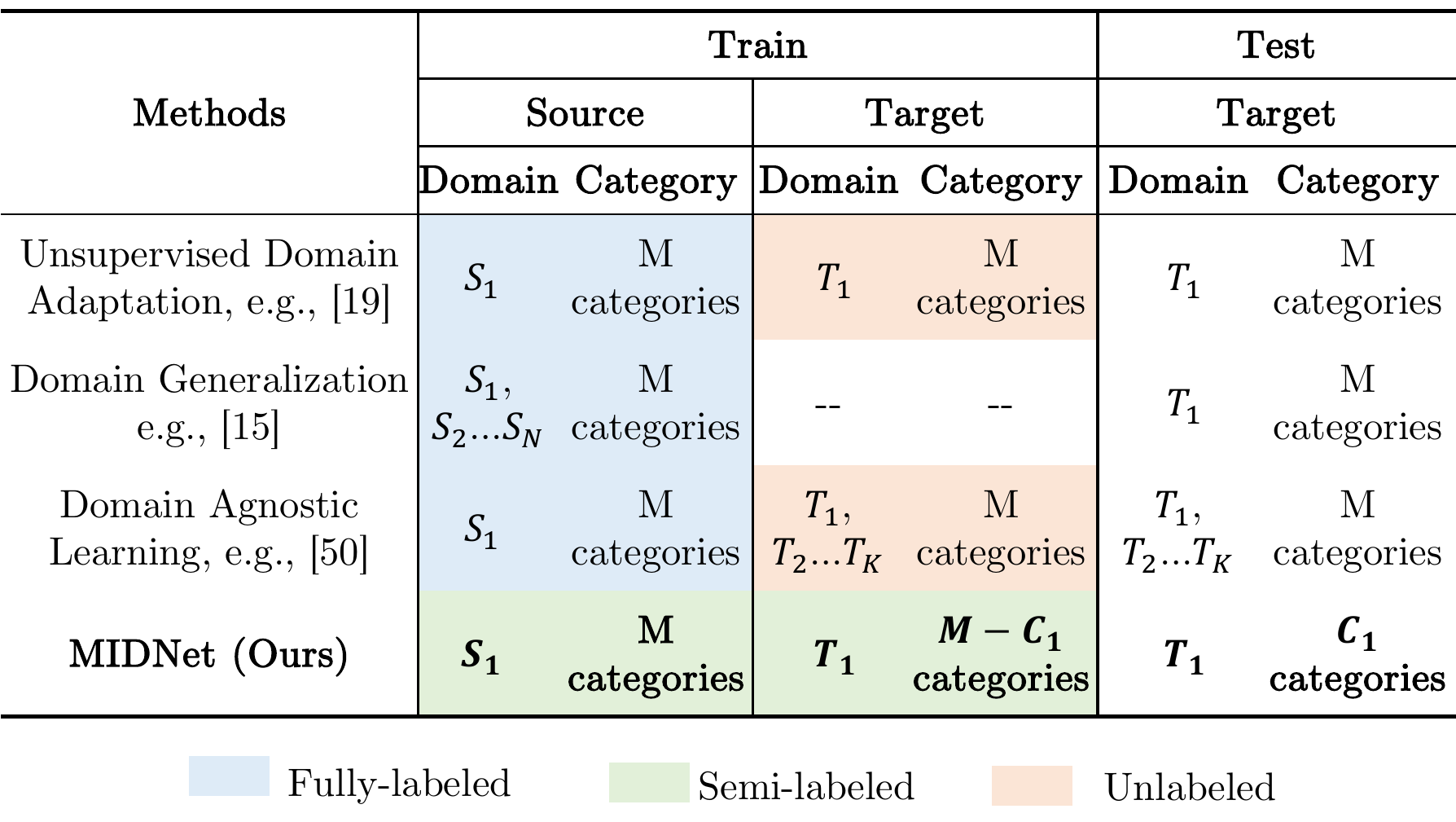}
 \caption{The differences between our method and other existing domain adaptation methods. We compare two aspects, the problem setup and the training paradigm in this taxonomy.}
 \label{DAcomparison}
\end{figure}

In this paper, we consider a one-to-one domain adaptation setting. In contrast to other works of one-to-one domain adaptation, in our work, the categories in the target domain are a subset of the categories in the source domain. Our ultimate goal is to learn categorical-discriminative knowledge from available categories in both domains to separate unseen categories in the target domain. 
Fig.~\ref{DAcomparison} compares the different task settings between our work and other domain adaptation works. 

\subsubsection{Transfer learning} 
Our problem setting also contains flavors of transfer learning, according to the nomenclature in~\cite{Pan2010b}. Transfer learning is a broader field that tackles domain shift and transfers knowledge between different datasets. Domain adaptation is a subcategory of transfer learning~\cite{Pan2010b}. Problem settings with different tasks between source and target domains are close to inductive transfer learning~\cite{Zadrozny04}. Problem settings considering domain shift between source and target domains within a single task are similar to domain adaptation or transductive transfer learning~\cite{Raina07}. Unsupervised transfer learning is a special case of inductive transfer learning, where only unlabeled data is available in both domains during training.
In this paper, we focus on transferring knowledge from a source domain to a target domain for a single task and tackle covariate shift (Sec.~\ref{intro}). Therefore, as suggested in~\cite{Pan2010b}, we frame our problem setting within the context of domain adaptation.

\section{Method}

Our goal is to disentangle categorical features from domain features to obtain generalizable features, so that our model can classify the categories in the target domain which have not been seen during training.
We formulate our task as follows: let $\mathcal{X}^S={\{\mathbf{x}_i^S\}}_{i=1}^{|C^S|}$ be the images from a source domain which contain categories $C^S$ and $\mathcal{X}^T={\{\mathbf{x}_i^T\}}_{i=1}^{|C^T|}$ be images from a target domain with categories $C^T,C^T \subset C^S$.
In both domains, categorical labels are available for part of the images as $\mathcal{Y}^S, \mathcal{Y}^T$. We want to train a network to maximize the categorical prediction performance of the classifier on images in the target domain from new categories ${\{\mathbf{x}^T|\mathbf{x}^T \in C^S-C^T\}}$. 

\begin{figure*}[tb]
 \centering
 \includegraphics[width=\textwidth, trim=8cm 10.3cm 8cm 3.1cm, clip]{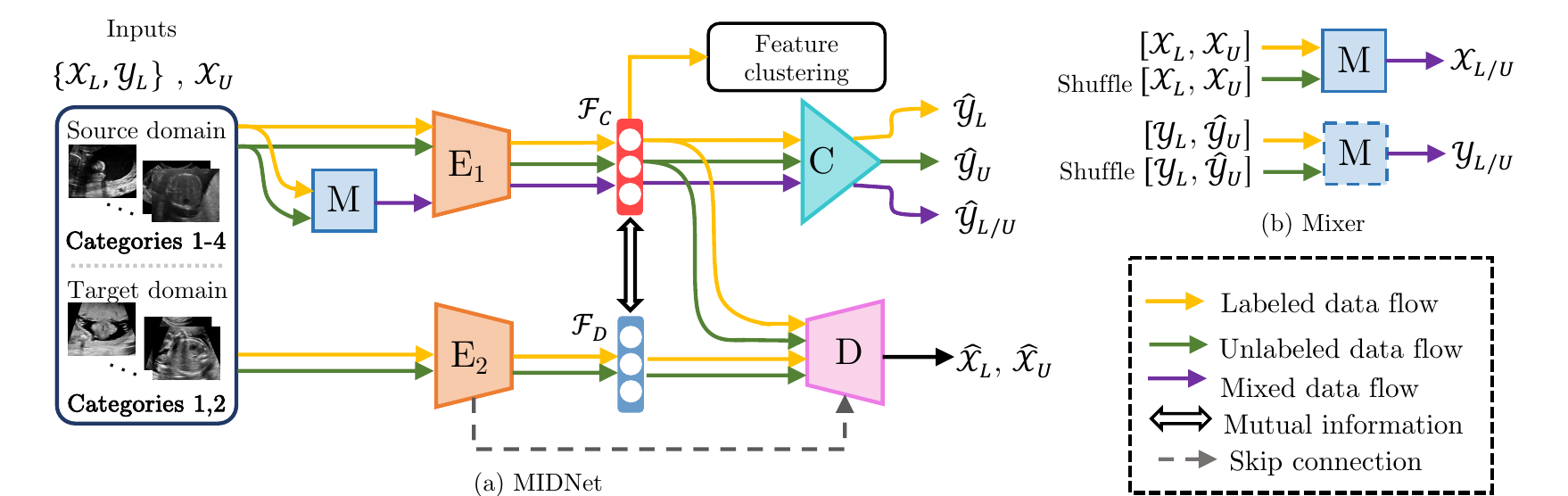}
 \caption{(a) An overview of MIDNet. We extract disjoint features ($\mathcal{F}_C, \mathcal{F}_D$) through mutual information minimization in the latent space and apply a feature consistency constraint to extract domain-invariant categorical features $\mathcal{F}_C$ for further disentanglement. $E_1, E_2$ are separate encoders, $C$ is a classifier and $D$ is a decoder. (b) We integrate unlabeled data by a mixer $M$ for semi-supervised learning.
 }
 \label{methed_outline}
\end{figure*}

To solve this task, we propose MIDNet in combination with semi-supervised learning. The architecture of our model is shown in Fig.~\ref{methed_outline}. Two independent encoders $E$ are utilized to respectively extract categorical features $\mathcal{F}_C$ and domain features $\mathcal{F}_D$ from labeled data $\{\mathcal{X}_L, \mathcal{Y}_L\}=\{\mathbf{x}_i|\mathbf{x}_i\in\mathcal{X}^S\cup\mathcal{X}^T, y_i|y_i\in\mathcal{Y}^S\cup\mathcal{Y}^T\}$ and unlabeled data $\mathcal{X}_U=\{\mathbf{x}_j|\mathbf{x}_j\in\mathcal{X}^S\cup\mathcal{X}^T\}$. The classifier $C$ is responsible for predicting class distributions from $\mathcal{F}_C$ for both $\mathcal{X}_L$ and $\mathcal{X}_U$ while the decoder $D$ combines $\mathcal{F}_C$ and $\mathcal{F}_D$ for the reconstruction of input images. The mixer $M$ aims to linearly mix labeled and unlabeled samples so that the model is trained to show linear behavior between samples for further leveraging of unlabeled data. For representation disentanglement, mutual information between $\mathcal{F}_C$ and $\mathcal{F}_D$ is minimized to encourage $\mathcal{F}_C$ to become domain-invariant and maximally informative for categorical classification. Feature clustering contains feature alignment and distance metric learning. Feature alignment aims at keeping the feature consistency between labeled images to promote the independence of $\mathcal{F}_C$. Distance metric learning considers inter-class relationships, which clusters similar samples while separating dissimilar samples to optimize $\mathcal{F}_C$ for improving classification performance. 

\subsection{Image reconstruction}

The first step of MIDNet is to employ an Encoder-Decoder framework for independent extraction of two internal representations from the input data $\mathbf{x}$. Two encoders $E_1, E_2$ are built to respectively generate latent vectors that aim to represent categorical features $\mathcal{F}_C$ and domain features $\mathcal{F}_D$, where $\mathcal{F}_C=E_1(\mathbf{x};\phi_1)$ and $\mathcal{F}_D=E_2(\mathbf{x};\phi_2)$. The decoder $D$ is utilized to guarantee that the combination of these features is capable of recovering original input data, where $\widehat{\mathbf{x}}=D(\mathcal{F}_C, \mathcal{F}_D;\psi)$. Here, $\mathbf{x}\in\mathcal{X}_L\cup\mathcal{X}_U$ and $\phi_1, \phi_2, \psi$ are the parameters of $E_1, E_2, D$, respectively. The cost function of this reconstruction is
\begin{equation}\label{Loss_reconstuction}
\mathcal{L}_{rec}=\| \widehat{x}-x\|_F^2.
\end{equation}

We concatenate layers between $E_2$ and $D$ to integrate high-frequency features from $E_2$ into the reconstruction, which helps $\mathcal{F}_D$ to contain valid information instead of noise.
This image reconstruction extracts two groups of features from internal representations of original data. 
The rest of our networks are designed and trained to enable $\mathcal{F}_C$ to only contain categorical information, thus becoming separated from $\mathcal{F}_D$ that only contains domain information.    
 
\subsection{Classification}
\label{cls}

We use a classifier $C$ to predict $|C^S|$ labels for labeled data, which encourages $\mathcal{F}_C$ to be maximally informative about categorical classification. $E_1, E_2$ and $C$ are updated by minimizing the cross-entropy loss 
\begin{equation}\label{Loss_classification}
\mathcal{L}_{cls}=-\mathbb{E}_{\{\mathbf{x},y\}\thicksim \{\mathcal{X}_L, \mathcal{Y}_L\}}\sum_{t=1}^{|C^S|}\mathbbm{1}[y=t]log(C(\mathcal{F}_C;\delta)).
\end{equation}
Here $\delta$ refers to the parameters of $C$. At the same time, $C$ predicts the class distribution of the unlabeled data $\mathbf{x} \in \mathcal{X}_U$ as $P_{C}(\widehat{y}|\mathbf{x};\delta), \widehat{y}\in\widehat{\mathcal{Y}}_U$. The predicted $P_{C}(\widehat{y}|\mathbf{x};\delta), \mathbf{x} \in \mathcal{X}_U$ will be utilized in SSL-based regularization (Sec.~\ref{mixmat}). The classifier on its own is unlikely to ensure that categorical features $\mathcal{F}_C$ are domain-invariant. This is because the training objective in Eq.~\ref{Loss_classification} only ensures that $\mathcal{F}_C$ contains as much information as possible for the target classification task.

\subsection{Mutual information disentanglement}
\label{mutual}

To address the problem from Sec.~\ref{cls}, we minimize the mutual information between $\mathcal{F}_C$ and $\mathcal{F}_D$. This minimization forces $\mathcal{F}_C$ to contain less domain information and thus separates categorical features from domain features.
Mutual information is defined as 
\begin{equation}\label{MI_orig}
I(\mathcal{D}_{\mathcal{F}_C};\mathcal{D}_{\mathcal{F}_D})=\int_{\mathcal{X}\times\mathcal{Z}}log\frac{d\mathbb{P}_{XZ}}{d\mathbb{P}_{X}\bigotimes\mathbb{P}_{Z}}d\mathbb{P}_{XZ}, 
\end{equation}
where $\mathbb{P}_{XZ}$ is the joint probability distribution of $(\mathcal{D}_{\mathcal{F}_C},\mathcal{D}_{\mathcal{F}_D})$, $\mathbb{P}_{X}=\int_\mathcal{Z}d\mathbb{P}_{XZ}$ and $\mathbb{P}_{Z}=\int_\mathcal{X}d\mathbb{P}_{XZ}$ are respectively marginal distributions of $\mathcal{D}_{\mathcal{F}_C}$ and $\mathcal{D}_{\mathcal{F}_D}$.
We utilize Mutual Information Neural Estimation (MINE)~\cite{Belghazi2018} to approximate the lower-bound of mutual information on $n$ samples by a neural network $T$ with parameters $\theta\in\Theta$, 
\begin{equation}\label{MINE}
\widehat{I(\mathcal{D}_{\mathcal{F}_C};\mathcal{D}_{\mathcal{F}_D})}_n=\sup_{\theta\in\Theta}E_{\mathbb{P}_{XZ}^{(n)}}\left[T_\theta\right]-log(E_{\mathbb{P}_{X}^{(n)}\bigotimes\widehat{\mathbb{P}}_{Z}^{(n)}}\left[e^{T_\theta}\right]).
\end{equation}
Practically, the expectations in Eq.~\ref{MINE} are estimated by Monte-Carlo integration~\cite{Peng2019} with shuffled samples along the batch axis ($\mathcal{F}'_D$), and thus the cost function of the mutual information disentanglement is 
\begin{equation}\label{MINE_shuffle}
\begin{split}
    \mathcal{L}_{MI}&=\widehat{I(\mathcal{D}_{\mathcal{F}_C};\mathcal{D}_{\mathcal{F}_D})}_n \\
    &=\frac{1}{n}\sum_{i=1}^nT(\mathcal{F}_C,\mathcal{F}_D,\theta)
    -log(\frac{1}{n}\sum_{i=1}^ne^{T(\mathcal{F}_C,\mathcal{F}'_D,\theta)}).
\end{split}
\end{equation}
Here, $(\mathcal{F}_C,\mathcal{F}_D)$ are sampled from joint distributions while $(\mathcal{F}_C,\mathcal{F}'_D)$ are sampled from the product of marginal distributions. 

\subsection{Feature clustering}
\label{FeaCluster}

To introduce the supervision from labeled samples for an enhanced disentanglement, we aim at aligning categorical features between source and target domains.
We hypothesize that categorical features of a certain category are supposed to be consistent between different domains. Therefore, we further enhance $\mathcal{F}_C$ to be domain-invariant by minimizing the distance of categorical features between source domain and target domain
for samples in $\mathcal{X}_L$ with
\begin{equation}\label{Loss_alignment}
\mathcal{L}_{align}=\frac{1}{|C^T|}\sum_{i=1}^{|C^T|}(\frac{1}{n_{c_i}}\sum_{j=1}^{n_{c_i}} \|f_{c_{ij}}^S - f_{c_{ij}}^T\|_F^2),
\end{equation}
where $n_{c_i}$ is the number of samples in category $c_i$, $c_i\in C_T$ and $f_{c_{ij}}^S, f_{c_{ij}}^T$ are the categorical features of the $j$th sample from category $c_i$ in the source domain and the target domain, respectively. 
This loss is computed within categorical features $\mathcal{F}_C$ in order to enhance mutual information disentanglement. This is different from feature alignment in other domain adaptation approaches (\emph{e.g.},~\cite{Sun2016,Long2017}), where the feature alignment directly aligns the whole latent features of inputs. In addition, our categorical feature alignment uses the Frobenius norm as the distance metric since labeled samples from both domains are available, whereas many other domain adaptation approaches (\emph{e.g.},~\cite{Tzeng2014,Long2015}) utilize Maximum Mean Discrepancy (MMD)~\cite{Borgwardt2006mmd} to diminish the discrepancy between a labeled source domain and an unlabeled target domain.

After feature alignment, We additionally consider inter-class relationships. To further cluster samples from the same category and separate samples from different categories in the latent space, we introduce distance metric learning with triplet loss~\cite{Florian2015} on $\mathcal{F}_C$ of labeled images from both domains,

\begin{equation}\label{triplet}
\mathcal{L}_{trip}=max\{0, d(f_{c_{iq}},f_{c_{ip}})-d(f_{c_{iq}},f_{c_{kn}})+\xi\}.
\end{equation}

Here, $d(\cdot,\cdot)$ is the squared Euclidean distance. $f_{c_{iq}}$ is the categorical feature of one query sample from category $c_i$. $f_{c_{ip}}$ and $f_{c_{kn}}$ are respectively categorical features of one support sample from the same category $c_i$ and one negative sample from a different category $c_k$, where $c_i, c_k\in C_S\cup C_T$. The selection of the query sample, the support sample and the negative sample is essential for converge. For implementation, we utilize the online semi-hard triplet mining strategy in~\cite{Florian2015}. 

Considering both feature alignment and separation, the feature clustering loss is
\begin{equation}\label{loss_clsutering}
\mathcal{L}_{clus}=\mathcal{L}_{align}+\eta\mathcal{L}_{trip}
\end{equation}
where $\eta$ is a hyper-parameter. In practice, we mainly focus on feature alignment and set $\eta=0$ as the proposed method because feature alignment is the key point to keep consistency between both domains. In our applications, we conduct both experiments,  \emph{i.e.}, with and without $\mathcal{L}_{trip}$, to evaluate the importance of inter-class relationships.   

\subsection{SSL-based regularization}
\label{mixmat}

Conventional fully supervised learning requires a large number of annotated input images with categorical labels and domain labels. However, in practice, labeled data is not easily available at any granularity. 
Berthelot et al.~\cite{Berthelot2019} propose a SSL-based method, MixMatch, integrating unlabeled data during training to reduce the dependency on labeled data. We adopt a simplified MixMatch using a mixer $M$ to leverage unlabeled data.
As shown in Fig.~\ref{methed_outline} (b), $M$ is utilized to linearly combine two random samples $(\mathbf{x}_1, \mathbf{x}_2)$ and their corresponding labels $(y_1, y_2)$ from labeled data and unlabeled data by 
\begin{equation}\label{mixUp}
\begin{split}
    &\mathbf{x}_{mix}=M(\mathbf{x}_1,\mathbf{x}_2;\beta)=\beta \mathbf{x}_1+(1-\beta)\mathbf{x}_2, \\ &y_{mix}=M(y_1,y_2;\beta)=\beta y_1+(1-\beta)y_2.
\end{split}
\end{equation}
where $\beta=max(\xi, 1-\xi)$, $\xi\thicksim Beta(\alpha,\alpha)$. Here, $\mathbf{x}_1\in\mathcal{X}_{cat}$, $\mathbf{x}_2\in\mathcal{X}'_{cat}$. $\mathcal{X}_{cat}=[\mathcal{X}_L,\mathcal{X}_U]$ is the concatenation of $\mathcal{X}_L$ and $\mathcal{X}_U$. $\mathcal{X}'_{cat}$ is the shuffled $\mathcal{X}_{cat}$ along the batch axis. Similarly, $y_1\in\mathcal{Y}_{cat}$ with $\mathcal{Y}_{cat}=[\mathcal{Y}_L,\widehat{\mathcal{Y}}_U]$, and $y_2\in\mathcal{Y}'_{cat}$. Note that $\widehat{\mathcal{Y}}_U$ is the collection of the predicted labels for unlabeled data according to Sec.~\ref{cls}. We denote that $\mathbf{x}_{mix}\in\mathcal{X}_{L/U}$, $y_{mix}\in\mathcal{Y}_{L/U}$. The goal of this SSL-based regularization is to encourage the linear behavior of the classifier, and thus the objective function is 
\begin{equation}\label{SSL}
\mathcal{L}_{SSL}=\|y_{mix}-P_C(\widehat{y}_{mix}|\mathbf{x}_{mix};\delta)\|_F^2, 
\end{equation}
where $P_C(\widehat{y}_{mix}|\mathbf{x}_{mix};\delta)=C(\mathbf{x}_{mix};\delta)$ is the predicted label of $\mathbf{x}_{mix}$ via classifier $C$.

\subsection{Optimization}

Our model is an end-to-end trainable framework and the overall objective is a linear combination of all cost functions 
\begin{equation}\label{Loss}
min\{\lambda_1\mathcal{L}_{rec}+\lambda_2\mathcal{L}_{cls}+\lambda_3\mathcal{L}_{MI}+\lambda_4\mathcal{L}_{clus}+\lambda_5\mathcal{L}_{SSL}\}, 
\end{equation}
where $\lambda_1$ to $\lambda_5$ are hyper-parameters chosen experimentally depending on the dataset. We optimize the MINE and the rest of our model in an alternating fashion. Inspired by~\cite{Belghazi2018}, we use the Adam optimizer ($\text{beta}=0.9$, $\text{learning rate}=10^{-5}$) to train the network parameters $\theta$ based on Eq.~\ref{MINE_shuffle} and use Stochastic Gradient Descent (SGD) with momentum optimizer ($\text{momentum}=0.9$, $\text{learning rate}=10^{-5}$) to update the parameters of encoders, decoders and classifier based on Eq.~\ref{Loss}.
We apply L2 regularization ($\text{scale}=10^{-5}$) to all weights during training to prevent over-fitting and we apply random image flipping as data augmentation. 
Classes are kept balanced on labeled data during training. Our model is trained on a Nvidia Titan X GPU with 12 GB of memory.

\section{Experiments}
We evaluate the proposed method on two fetal US standard plane classification tasks, where the domain shifts are respectively caused by shadow artifacts (Fig.~\ref{DataPresentation}(a)) and different image acquisition devices (Fig.~\ref{DataPresentation}(b)). For both tasks, images from source and target domains are unpaired and collected independently. 
We show the key results in the main paper and detailed implementation, network architectures as well as more results in the supplemental Appendix.  

\subsection{Experiment settings}
We conduct comprehensive evaluations for each application in the following. (1) We compare the proposed method with the state-of-the-art algorithms for domain adaptation. (2) We explore the effectiveness of different key components in MIDNet via an ablation study. (3) To demonstrate the importance of unlabeled data, we compare the classification performance with and without unlabeled training data on the target domain. (4) By training MIDNet with different percentage of labeled data, we evaluate the performance of our model in a  semi-supervised setting. In addition, we discuss the influence of common categories on the classification performance of unseen categories in the target domain.

We utilize three groups of test data for the evaluation: (\textit{i}) test data from the source domain $T_{Source}$, 
(\textit{ii}) test data from the target domain whose image features have been observed during training $T_{Target}$, 
and (\textit{iii}), most importantly, test data from the target domain whose image features are absent during training $T_{Target}^{New}$.

We adopt commonly-used statistical metrics, F1-score, recall and precision, to quantitatively evaluate classification performance.
$\text{Recall}=TP/(TP+FN)$, $\text{Precision}=TP/(TP+FP)$ and F1-score is the the harmonic mean of precision and recall. We report the average scores of these metrics for all examined methods. As suggested by~\cite{Shai2007}, we utilize the $\mathcal{A}$-distance as a measure of domain divergence to quantitatively evaluate the separation of categorical and domain features. Similar to~\cite{Ganin2016, Saito2019}, we train a SVM as a domain classifier to compute $\epsilon$ (the error of classifier) for the $\mathcal{A}$-distance, $\hat{d}_{\mathcal{A}}=2(1-2\epsilon)$. 

\subsection{Comparison methods and ablation study}
\label{comparison_ablation}
We evaluate a VGG network~\cite{Simonyan15} which is trained on data only from the source domain, namely \textit{Source only}, as a baseline to demonstrate that the domain shift problems affects the generalizability of deep models. To verify that MIDNet is able to extract generalized features across domains, we compare MIDNet with a VGG network~\cite{Simonyan15} and
a VGG network with residual unit~\cite{He2016} (Res-VGG). We further compare MIDNet to the state-of-the-art feature disentanglement algorithms for addressing the task in this work, including a two-step disentanglement method\footnote{https://github.com/naamahadad/A-Two-Step-Disentanglement-Method}~\cite{Hadad2018} and a multi-task learning based disentanglement method\footnote{https://github.com/qmeng99/Multi-task-Representation-Disentanglement}~\cite{meng2019}. Note that we implement the method in~\cite{Hadad2018} differently from the original paper. Specifically, we train the model simultaneously to enable it to be suitable for our task setup. We denote \textit{Two-step-fair} as~\cite{Hadad2018} with an adversarial network using unspecific features ($Z$) for category classification and denote \textit{Two-step-Unfair} as~\cite{Hadad2018} with an adversarial network using specific features ($S$) for domain classification.  
We keep the original experimental settings for the method in~\cite{meng2019} (namely Multi-task). All comparison methods above are fully-supervised. Additionally, we compare the proposed method with the state-of-the-art domain adaptation methods, including domain-adversarial training of neural networks (DANN)\footnote{https://github.com/pumpikano/tf-dann, Jan 2018}~\cite{Ganin2016} and semi-supervised domain adaptation via MiniMax Entropy (MME)\footnote{https://github.com/VisionLearningGroup/SSDA\_MME}~\cite{Saito2019}. These two comparison methods are semi-supervised. For all comparison methods with hyperparameters, we run several sets of parameter values (including the values in corresponding papers). The hyperparameters with the best experimental results are selected for the following evaluation.

For the ablation study, we remove different loss components to obtain different combinations of components in MIDNet. The detailed combinations of ablations are shown in Table.~\ref{ablation_table}.

\begin{table}[tb]
\centering
\caption{Different combination of key components in MIDNet for the ablation study. $\mathcal{L}_{rec}$ is for reconstruction, $\mathcal{L}_{cls}$ is for classification, $\mathcal{L}_{MI}$ is for mutual information disentanglement and $\mathcal{L}_{SSL}$ is for integrating unlabeled data. For $\mathcal{L}_{clus}$, $\eta=0$ only considers feature alignment and $\eta\neq0$ considers both feature alignment and inter-class relationships.}
\label{ablation_table}
\begin{threeparttable}
\begin{tabular}{l|cccccc}
\toprule[1.2pt]
Methods          &
$\mathcal{L}_{rec}$    & 
$\mathcal{L}_{cls}$    & 
$\mathcal{L}_{MI}$     &
\thead{$\mathcal{L}_{clus}$ \\ $\eta=0$} & 
$\mathcal{L}_{SSL}$    & 
\thead{$\mathcal{L}_{clus}$ \\ $\eta\neq0$}      \\
\midrule
MIDNet-I               & 
$\surd$                 & 
$\surd$                  &
~~~                    &
~~~                    &
~~~                    &
~~~                    \\
MIDNet-II               & 
$\surd$                  & 
$\surd$                  &
$\surd$                    &
~~~                    &
~~~                    &
~~~                    \\
MIDNet-III               & 
$\surd$                  & 
$\surd$                  &
~~~                    &
$\surd$                    &
~~~                    &
~~~                    \\
MIDNet-IV               & 
$\surd$                  & 
$\surd$                  &
~~~                    &
~~~                    &
$\surd$                    &
~~~                    \\
MIDNet w/o $\mathcal{L}_{SSL}$               & 
$\surd$                  & 
$\surd$                  &
$\surd$                    &
$\surd$                   &
~~~                    &
~~~                    \\
MIDNet w/o $\mathcal{L}_{clus}$               & 
$\surd$                  & 
$\surd$                  &
$\surd$                    &
~~~                    &
$\surd$                    &
~~~                    \\
MIDNet w/o $\mathcal{L}_{MI}$               & 
$\surd$                  & 
$\surd$                  &
~~~                    &
$\surd$                    &
$\surd$                    &
~~~                    \\                      
MIDNet               & 
$\surd$                  & 
$\surd$                  &
$\surd$                    &
$\surd$                    &
$\surd$                    &
~~~                    \\
MIDNet+$\mathcal{L}_{trip}$               & 
$\surd$                  & 
$\surd$                  &
$\surd$                    &
~~~                    &
$\surd$                    &
$\surd$                    \\
\bottomrule[1.2pt]
\end{tabular}
\begin{tablenotes}
\item $\eta$ is experimentally selected for different tasks when $\eta\neq0$.
\end{tablenotes}
\end{threeparttable}
\end{table}

\subsection{Experiments on fetal US with and without shadows}
\label{NS_S_experiment}
The fetal US dataset consists of $\sim 7k$ 2D fetal US images sampled from 2694 2D US examinations with gestational ages between $18-22$ weeks (iFIND Project~\footnote{http://www.ifindproject.com/ \label{ifindfoot}}). Eight different US systems of identical make and model (GE Voluson E8) were used for the acquisitions to eliminate as many unknown image acquisition parameters as possible. Six different anatomical standard plane locations have been selected by an experienced sonographer, including \emph{Four Chamber View (4CH)}, \emph{Abdominal}, \emph{Femur}, \emph{Lips}, \emph{Left Ventricular Outflow Tract (LVOT)} and \emph{Right Ventricular Outflow Tract (RVOT)}. 
The images have additionally been classified by an expert observer as shadow-containing or shadow-free.
In this experiment, the source domain contains shadow-free images (see Fig.~\ref{DataPresentation} (b) SF) while the target domain has shadow-containing images from less favorable imaging conditions (see Fig.~\ref{DataPresentation} (b) SC). Training data consists of all six standard planes from the source domain as well as Abdominal, LVOT and RVOT from the target domain. We aim to separate anatomical features (categorical features) and shadow artifacts features (domain features) to obtain generalized anatomical features for achieving high performance of standard plane classification on $T_{Target}^{New}$ (4CH, Femur and Lips from target domain). Here, $T_{Source}$ contains 4CH, Abdominal, Femur, Lips, LVOT and RVOT from the source domain and $T_{Target}$ contains Abdominal, LVOT and RVOT from the target domain.
Hyper-parameters $\lambda_1$ to $\lambda_5$ in Eq.~\ref{Loss} are $\lambda_1=1, \lambda_2=10, \lambda_3=10^{-4}, \lambda_4=10, \lambda_5=10$ for the proposed MIDNet model and $ \eta=0.005$ is additionally for MIDNet+$\mathcal{L}_{trip}$. 

\subsubsection*{Results} 
\begin{figure}
    \centering
    \subfloat[][Source: shadow-free;
    
    \hspace*{3.3em} Target: shadow-containing]{
    \includegraphics[height=3.5cm]{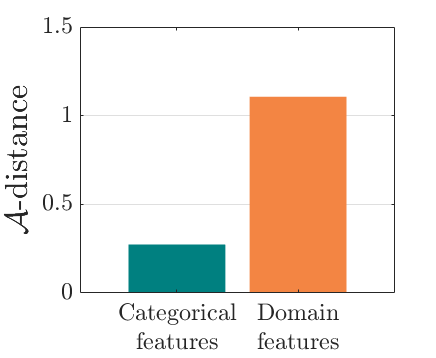}
    }
    \subfloat[][Source: device A; 
    
    \hspace*{1em}Target: device B]{
    \includegraphics[height=3.5cm]{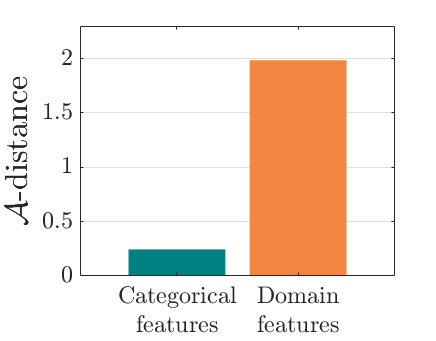}
    }
    \caption{Domain discrepancy based on categorical features and domain features in MIDNet. A high $\mathcal{A}$-distance score means a high domain difference.}
    \label{Adist}
\end{figure}

\begin{figure}
    \centering
    \subfloat[]{\includegraphics[height=3cm, trim=2cm 1.5cm 1.5cm 1.5cm, clip]{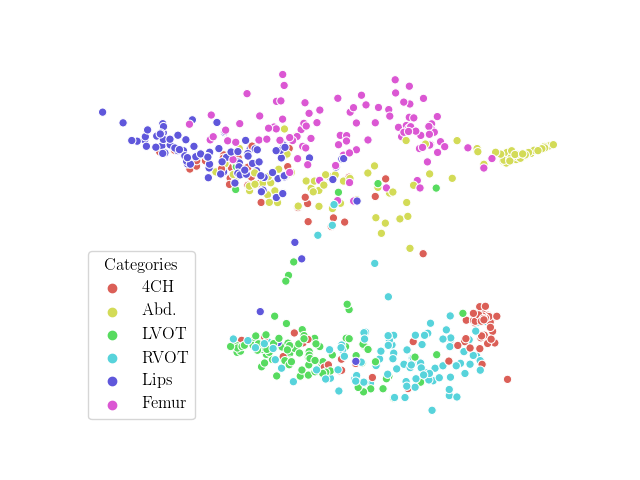}
    }
    \subfloat[]{
    \includegraphics[height=3cm, trim=2cm 1.5cm 1.5cm 1.5cm, clip]{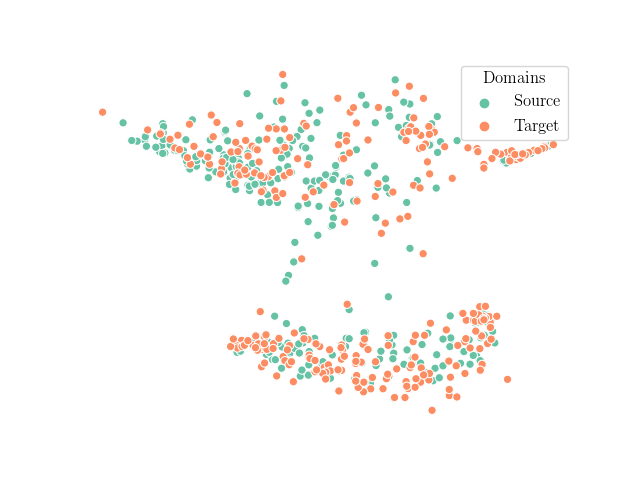}
    }
    \caption{Feature visualization of MIDNet with t-SNE. We plot categorical features from both domains. (a) The color represents categories. (b) The color represents domains. We observe that the categorical features  are domain-invariant and enable the anatomical classification.}
    \label{tsne_sns}
\end{figure}

We compare the $\mathcal{A}$-distance of categorical features and domain features. Fig.~\ref{Adist} (a) shows that domain difference is higher in domain features than in categorical features. This indicates that domain features contain more domain information whereas categorical features are more domain-invariant. Fig.~\ref{tsne_sns} shows the t-SNE plot of categorical features in both domains for MIDNet. From Fig.~\ref{tsne_sns} (a), we observe that the categorical features learned by MIDNet enable the anatomical classification. Fig.~\ref{tsne_sns} (b) shows that the learned categorical features are domain-invariant.

\begin{figure}[t]
    \centering
    \subfloat[][Source: shadow-free;
    
    \hspace*{3.3em} Target: shadow-containing]{
    \includegraphics[height=3.3cm]{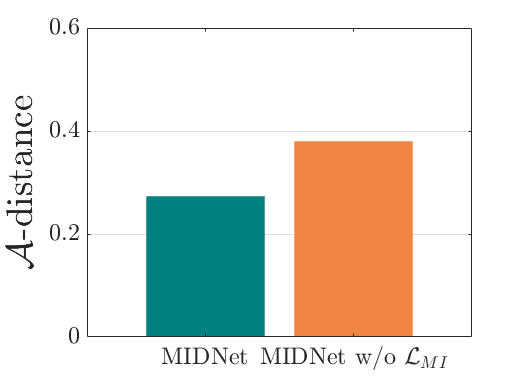}
    }
    \subfloat[][Source: device A; 
    
    \hspace*{1em}Target: device B]{
    \includegraphics[height=3.3cm]{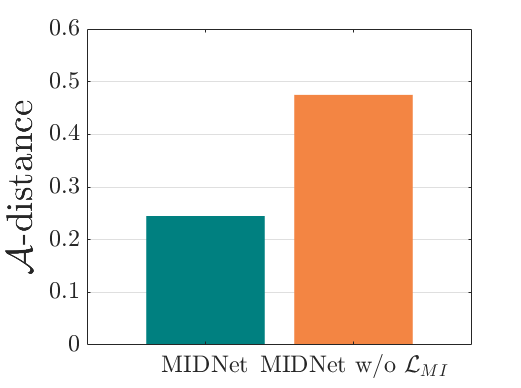}
    }
    \caption{Domain discrepancy of MIDNet and MIDNet w/o $\mathcal{L}_{MI}$ based on categorical features. A high $\mathcal{A}$-distance score means a high domain difference. Experiment setting is the same with Table~\ref{US_table_shadow}.}
    \label{Adist_lmi}
\end{figure}

The experimental results of the state-of-the-art and the ablation study are shown in Table.~\ref{US_table_shadow}. In this table, we observe that the MIDNet model outperforms all the state-of-the-art methods on the most important test data $T_{Target}^{New}$ for average F1-score, recall and precision. MIDNet+$\mathcal{L}_{trip}$ performs better than MIDNet on $T_{Source}$ and $T_{Target}$, demonstrating that metric learning is important and efficient for improving classification performance on images whose features have been observed during training. In the ablation study, MIDNet outperforms other variant models, especially \textit{MIDNet w/o $\mathcal{L}_{SSL}$}, \textit{MIDNet w/o $\mathcal{L}_{clus}$} and \textit{MIDNet w/o $\mathcal{L}_{MI}$}, illustrating the effectiveness of all proposed components in MIDNet. In addition, Fig.~\ref{Adist_lmi} (a) shows that the $\mathcal{A}$-distance of \textit{MIDNet w/o $\mathcal{L}_{MI}$} is higher than that of MIDNet. This demonstrates that mutual information disentanglement ($\mathcal{L}_{MI}$) contributes to learn domain-invariant categorical features. 

\begin{figure}[ptb]
 \centering
 \subfloat[Source: shadow-free; Target: shadow-containing]{
 \begin{tabular}{@{\hspace{-2\tabcolsep}}c@{\hspace{0.3\tabcolsep}}c}
  \raisebox{0.5\height}{\rotatebox[origin=c]{0}{\makecell{~\scalebox{0.8}{\textbf{With unlabeled data}}}}}  &
  \raisebox{0.5\height}{\rotatebox[origin=c]{0}{\makecell{~\scalebox{0.8}{\textbf{Without unlabeled data}}}}}  \\
  \includegraphics[height=3.8cm, trim=2.2cm 0cm 3cm 1cm, clip]{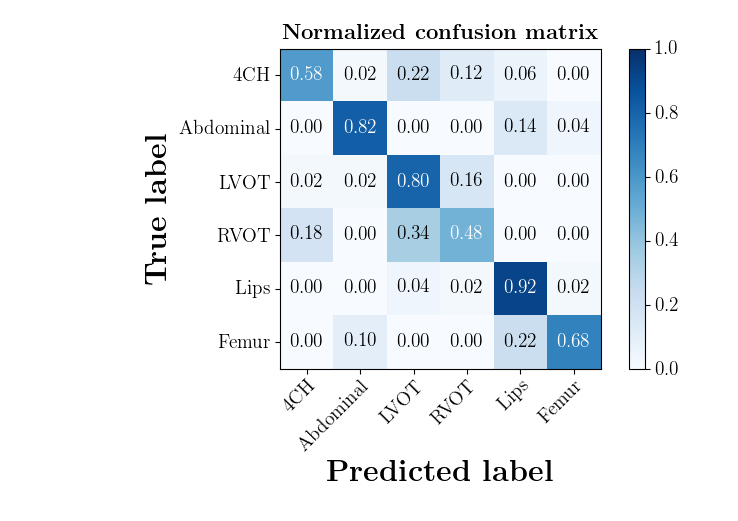} &
  \includegraphics[height=3.8cm, trim=2.2cm 0cm 1cm 1cm, clip]{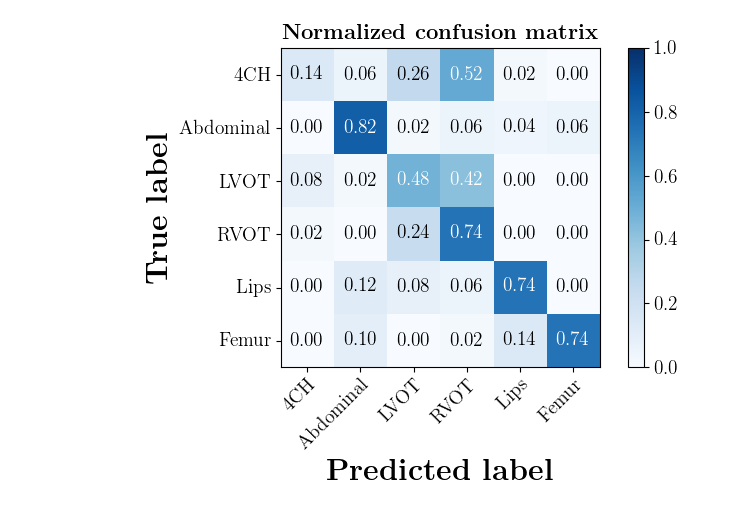}
  \end{tabular}
  } \\
 \subfloat[Source: device A; Target: device B]{
 \begin{tabular}{@{\hspace{-2\tabcolsep}}c@{\hspace{0.3\tabcolsep}}c}
 \raisebox{0.5\height}{\rotatebox[origin=c]{0}{\makecell{~\scalebox{0.8}{\textbf{With unlabeled data}}}}}  &
 \raisebox{0.5\height}{\rotatebox[origin=c]{0}{\makecell{~\scalebox{0.8}{\textbf{Without unlabeled data}}}}}  \\
  \includegraphics[height=3.8cm, trim=2.2cm 0cm 3cm 1cm, clip]{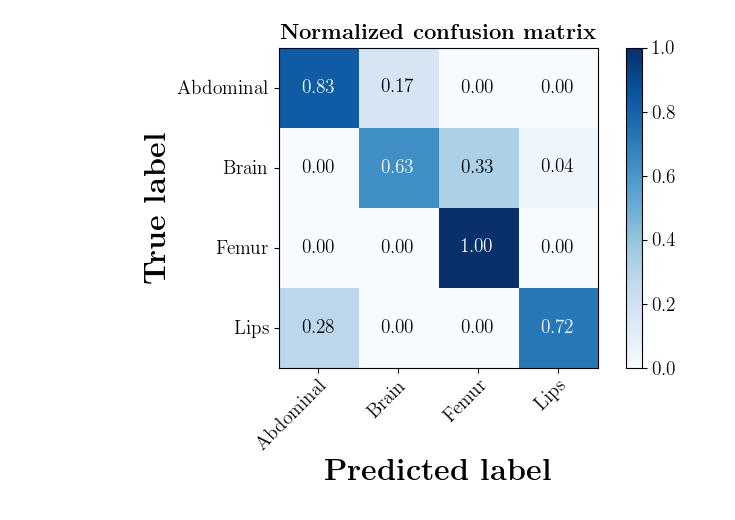} &
  \includegraphics[height=3.8cm, trim=2.2cm 0cm 1.2cm 1cm, clip]{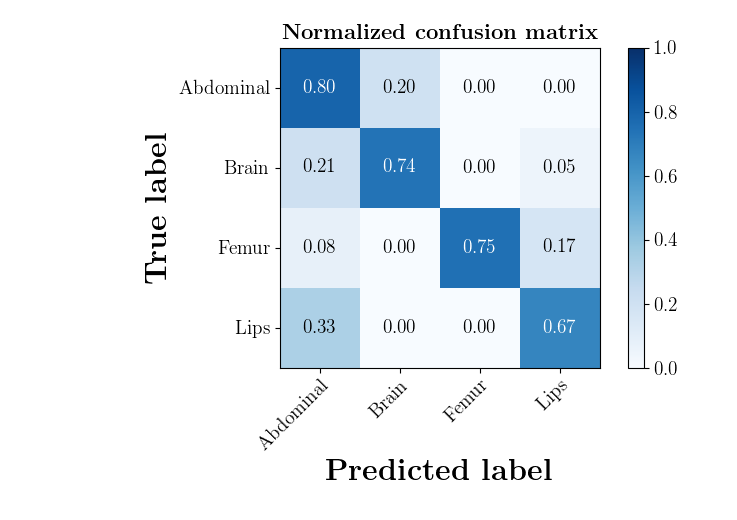}
  \end{tabular}
  }
  \caption{Confusion matrices of $\mathbf{T_{Target}}$ $\&$ $\mathbf{T_{Target}^{New}}$ in MIDNet with and without unlabeled data. For the \textit{with unlabeled data} setting, $30\%$ of training data are labeled and the rest are unlabeled. The \textit{without unlabeled data} setting only uses the $30\%$ labeled data for training, without using unlabeled data. (a) Classification on fetal US with/without shadows. (b) Classification on fetal US from different acquisition devices.}
  \label{CM_MIDNet}
\end{figure}

We further compare the performance of MIDNet with and without unlabeled data on the target domain. Here, the \textit{with unlabeled data} setting utilizes the training data containing $30\%$ labeled data and $70\%$ unlabeled data, while the \textit{without unlabeled data} setting only uses the $30\%$ labeled data. The confusion matrices in Fig.~\ref{CM_MIDNet}(a) show the effectiveness of unlabeled data in the proposed method, for example, the classification accuracy of $T_{Target}^{New}$ in MIDNet (e.g., \textit{4CH} and \textit{Lips}) improves when integrating unlabeled data.

\begin{figure*}[tb]
 \centering
 \subfloat[Source: shadow-free; Target: shadow-containing]{
  \includegraphics[width=\columnwidth, trim=1.2cm 0cm 1.1cm 0.2cm, clip]{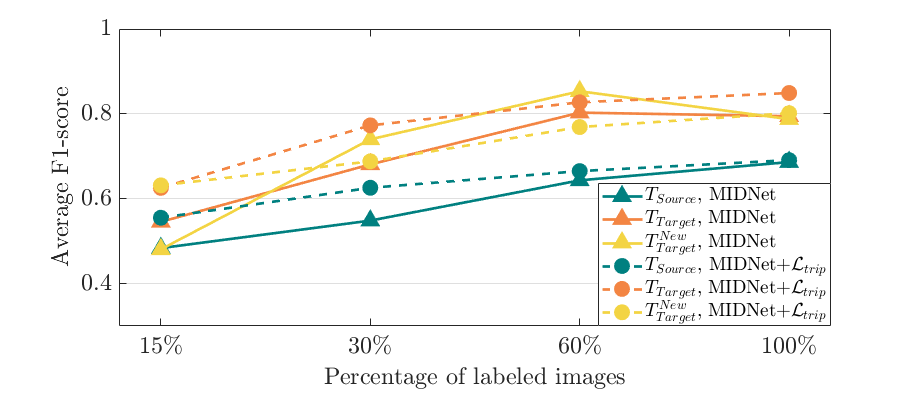}
  } 
  \subfloat[Source: device A; Target: device B]{
  \includegraphics[width=\columnwidth, trim=1.2cm 0cm 1.1cm 0.2cm, clip]{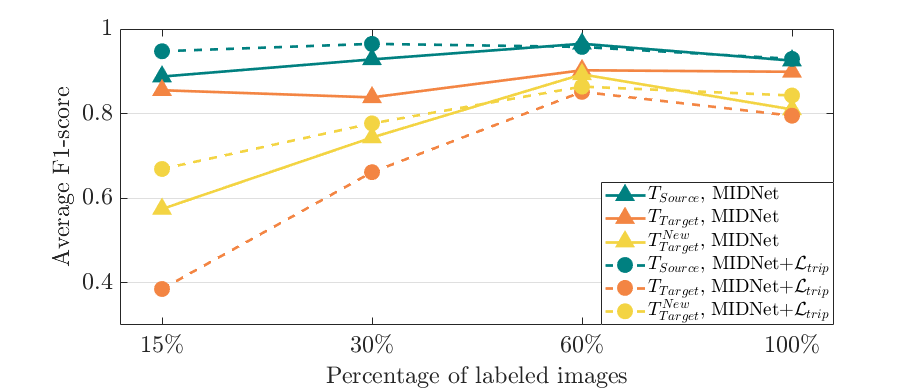}
  } 
  \caption{Average F1-score of standard plane classification with different percentage of labeled data ($15\%, 30\%, 60\%, 100\%$) for semi-supervised learning based on MIDNet model and MIDNet+$\mathcal{L}_{trip}$. (a) Classification on fetal US with/without shadows. (b) Classification on fetal US from different image acquisition devices.}
  \label{semi}
\end{figure*}

To explore the importance of labeled data, we evaluate the performance of MIDNet and MIDNet+$\mathcal{L}_{trip}$ based on using $15\%, 30\%, 60\%$ and $100\%$ labeled data during training. Fig.~\ref{semi} (a) shows the average F1-score on three groups of test data. From this figure, we observe that the classification performance improves with increasing labeled data.

Finally, correctly classified and mis-classified examples of $T_{Target}^{New}$ using MIDNet are presented in Fig.~\ref{TPFP_ex} (a).

\begin{table*}[tb]
\centering
\caption{Comparison of the state-of-the-art and ablation study for the \textbf{fetal US standard plane classification task using data with/without shadow artifacts}. $30\%$ of training data are labeled data and the rest are unlabeled data. Average F1-score, Recall and Precision are measured on three groups of test data. Best results are shown in bold.}
\label{US_table_shadow}
\resizebox{\textwidth}{!}{
\begin{threeparttable}
\begin{tabular}{lccc|ccc|ccc}
\toprule[1.2pt]
\multirow{2}{*}{Methods}                            & 
\multicolumn{3}{c|}{$T_{Source}$}                    &
\multicolumn{3}{c|}{$T_{Target}$}                    &
\multicolumn{3}{c}{$T_{Target}^{New}$}              \\
\cmidrule{2-10}
~~~~~                                               &
F1-score                                                & 
Recall                                              & 
Precision                                           &
F1-score                                                & 
Recall                                              & 
Precision                                           &
F1-score                                                & 
Recall                                              & 
Precision                                           \\
\midrule
Source only                                   &
0.4878                                              &
0.4933                                              &
0.4914                                              &
0.4341                                              &
0.4067                                              &
0.4693                                              &
0.5496                                              &
0.5867                                              &
0.5170                                              \\
VGG~\cite{Simonyan15}                                                 &
0.5347                                              &
0.5400                                              &
0.5335                                              &
0.4492                                              &
0.4667                                              &
0.4337                                              &
0.5495                                              &
0.5400                                              &
0.5614                                              \\
Res-VGG~\cite{Simonyan15,He2016}                    &
0.4669                                              &
0.4800                                              &
0.5229                                              &
0.5482                                              &
0.5867                                              &
0.5293                                              &
0.5722                                              &
0.5467                                              &
0.6182                                              \\
Two-step-fair~\cite{Hadad2018}                                       &
0.4531                                              &
0.4500                                              &
0.4572                                             &
0.4400                                              &
0.4467                                              &
0.4338                                              &
0.5008                                              &
0.4933                                              &
0.5095                                              \\
Two-step-Unfair~\cite{Hadad2018}                                       &
0.4894                                              &
0.4933                                              &
0.4895                                             &
0.4515                                              &
0.4733                                              &
0.4319                                              &
0.4571                                              &
0.4400                                              &
0.4769                                              \\
Multi-task~\cite{meng2019}                                          &
0.4622                                              &
0.4667                                              &
0.5524                                             &
0.5787                                              &
0.5667                                              &
0.6220                                              &
0.6393                                              &
0.6533                                              &
0.6491                                              \\
DANN~\cite{Ganin2016}                        &
0.5939                                              &
0.6100                                              &
0.6598                                              &
0.6642                                              &
0.7000                                              &
0.6400                                              &
0.5525                                              &
0.5533                                              &
0.5700                                              \\
MME~\cite{Saito2019}                        &
0.5723                                              &
0.5700                                              &
\textbf{0.7496}                                              &
\textbf{0.8163}                                              &
0.8000                                              &
\textbf{0.8458}                                              &
0.4852                                              &
0.5200                                              &
0.5006                                              \\
\midrule
MIDNet-I                                           &
0.4643                                              &
0.4767                                              &
0.5891                                             &
0.5796                                              &
0.5933                                              &
0.5944                                              &
0.6280                                              &
0.6133                                              &
0.6947                                              \\
MIDNet-II                                           &
0.4760                                              &
0.4867                                              &
0.5336                                             &
0.6185                                              &
0.6533                                              &
0.6056                                              &
0.6559                                              &
0.6200                                              &
0.7412                                              \\
MIDNet-III                                           &
0.4929                                              &
0.5100                                              &
0.5498                                             &
0.5620                                              &
0.5800                                              &
0.5512                                              &
0.6887                                              &
0.6667                                              &
0.7267                                              \\
MIDNet-IV                                           &
0.4636                                              &
0.4833                                              &
0.5403                                             &
0.5746                                              &
0.5867                                              &
0.5705                                              &
0.6378                                              &
0.6400                                              &
0.6732                                              \\
MIDNet w/o $\mathcal{L}_{SSL}$                      &
0.5379                                              &
0.5533                                              &
0.6007                                             &
0.5976                                              &
0.6600                                              &
0.5612                                              &
0.6603                                              &
0.6000                                              &
0.8119                                              \\
MIDNet w/o $\mathcal{L}_{clus}$                      &
0.4195                                              &
0.4367                                              &
0.5102                                             &
0.5657                                              &
0.5800                                              &
0.5637                                              &
0.6025                                              &
0.6000                                              &
0.6539                                              \\
MIDNet w/o $\mathcal{L}_{MI}$                     &
0.5339                                              &
0.5467                                              &
0.5948                                             &
0.6654                                              &
0.7067                                              &
0.6449                                              &
0.7091                                              &
0.6600                                              &
0.8255                                              \\
MIDNet                       &
0.5484                                              &
0.5667                                              &
0.6683                                             &
0.6809                                              &
0.7000                                              &
0.6803                                              &
\textbf{0.7399}                                              &
\textbf{0.7267}                                              &
0.7830                                              \\
MIDNet+$\mathcal{L}_{trip}$                        &
\textbf{0.6257}                                              &
\textbf{0.6367}                                              &
0.7082                                              &
0.7728                                              &
\textbf{0.8533}                                              &
0.7146                                              &
0.6880                                              &
0.6200                                              &
\textbf{0.8396}                                              \\
\bottomrule[1.2pt]
\end{tabular}
\begin{tablenotes}
\item The baselines and ablation study models are introduced in Sec.~\ref{comparison_ablation}. DANN, MME and the proposed models are semi-supervised while other comparison methods are fully-supervised which only use the $30\%$ labeled images.
\end{tablenotes}
\end{threeparttable}
}
\end{table*}

\subsection{Experiments on cross-device fetal US}
\label{crossD_exp}
The previous experiment on fetal US images is supported by data restricted to one US imaging device. Here, we evaluate MIDNet for a standard plane classification task on data from different imaging devices (different device domains). 
Device A is ``GE Voluson E8" which is the same device in Sec.~\ref{NS_S_experiment}, which acquired $\sim6K$ 2D fetal US images. Device B is ``Philips EPIQ V7 G" which acquired another $\sim5K$ images sampled from about 500 2D US examinations with gestational ages between 20-32 weeks (see Fig.~\ref{DataPresentation} (b)). In this experiment, we use four different anatomical standard plane locations with sufficient images in both domains, including \emph{Abdominal}, \emph{Brain}, \emph{Femur} and \emph{Lips}, which are selected by an 10-year-experienced sonographer.
In this experiment, the source domain is set as device A while the target domain is device B.
Training data consists of all four standard planes from the source domain as well as Abdominal and Brain from the target domain. We aim to separate anatomical features (categorical features) and imaging device features (domain features) to obtain generalized anatomical features for achieving high performance of standard plane classification on $T_{Target}^{New}$ (Femur and Lips). Here, $T_{Source}$ contains Abdominal, Brain, Femur and Lips from the source domain and $T_{Target}$ contains Abdominal and Brain from the target domain.
Hyper-parameters $\lambda_1$ to $\lambda_5$ in Eq.~\ref{Loss} are $\lambda_1=1, \lambda_2=10, \lambda_3=10^{-4}, \lambda_4=50, \lambda_5=50$ for the proposed MIDNet model and $ \eta=2\times10^{-4}$ is additionally for MIDNet+$\mathcal{L}_{trip}$.

\begin{table*}[htb]
\centering
\caption{Comparison of \textit{Source only}, the state-of-the-art methods and ablation study for the \textbf{fetal US standard plane classification task with data from different acquisition devices (Source domain: Device A, Target domain: Device B)}. $30\%$ of training data are labeled data and the rest are unlabeled data. Best results in bold.}
\label{US_table_device_A2B}
\resizebox{\textwidth}{!}{
\begin{tabular}{lccc|ccc|ccc}
\toprule[1.2pt]
\multirow{2}{*}{Methods}                            & 
\multicolumn{3}{c|}{$T_{Source}$}                    &
\multicolumn{3}{c|}{$T_{Target}$}                    &
\multicolumn{3}{c}{$T_{Target}^{New}$}              \\
\cmidrule{2-10}
~~~~~                                               &
F1-score                                                & 
Recall                                              & 
Precision                                           &
F1-score                                                & 
Recall                                              & 
Precision                                           &
F1-score                                                & 
Recall                                              & 
Precision                                           \\
\midrule
Source only                                   &
0.7665                                              &
0.7700                                              &
0.7656                                              &
0.6971                                              &
0.6750                                              &
0.7305                                              &
0.6742                                              &
0.7050                                              &
0.6899                                              \\
VGG~\cite{Simonyan15}                           &
0.7565                                              &
0.7600                                              &
0.7553                                              &
0.6853                                              &
0.6750                                              &
0.6964                                              &
0.7039                                              &
0.7250                                              &
0.7011                                              \\
Res-VGG~\cite{Simonyan15,He2016}                    &
0.9196                                              &
0.9200                                              &
0.9229                                              &
0.5582                                              &
0.6200                                              &
0.5092                                              &
0.6880                                              &
0.6300                                              &
0.8728                                              \\
Two-step-fair~\cite{Hadad2018}                   &
0.7979                                              &
0.7975                                              &
0.8007                                              &
0.3965                                              &
0.4050                                              &
0.3919                                              &
0.7491                                              &
0.7400                                              &
0.7644                                              \\
Two-step-Unfair~\cite{Hadad2018}                &
0.7899                                              &
0.7900                                              &
0.7903                                              &
0.3869                                              &
0.4050                                              &
0.3831                                              &
0.6069                                              &
0.6150                                              &
0.6013                                              \\
Multi-task~\cite{meng2019}                        &
0.8124                                              &
0.8150                                              &
0.8383                                              &
0.3964                                              &
0.4200                                              &
0.4186                                              &
0.7522                                              &
0.7800                                              &
0.7955                                              \\
DANN~\cite{Ganin2016}                            &
0.9572                                  &
0.9575                                  &
0.9588                                  &
0.5507                                              &
0.6100                                              &
0.5521                                              &
0.5611                                              &
0.5050                                             &
0.7784                                              \\
MME~\cite{Saito2019}                                &
0.9526                                              &
0.9525                                              &
0.9537                                              &
0.5400                                          &
0.7150                                               &
0.4345                                              &
0.4293                                              &
0.3600                                             &
0.9595                                              \\
\midrule
MIDNet-I                                           &
0.9623                                              &
0.9625                                              &
0.9642                                              &
0.5572                                              &
0.5600                                              &
0.6256                                              &
0.3236                                              &
0.5000                                              &
0.2393                                              \\
MIDNet-II                                           &
0.9520                                              &
0.9525                                              &
0.9592                                              &
0.6665                                              &
0.6600                                              &
0.6731                                              &
0.3139                                              &
0.3950                                              &
0.6811                                              \\
MIDNet-III                                           &
0.8948                                              &
0.8950                                              &
0.8948                                              &
0.7125                                              &
0.8000                                              &
0.7383                                              &
0.7400                                              &
0.6750                                              &
\textbf{0.9630}                                              \\
MIDNet-IV                                           &
0.9446                                              &
0.9450                                              &
0.9486                                              &
0.1992                                              &
0.1300                                              &
0.4261                                              &
0.2473                                              &
0.5000                                              &
0.1471                                              \\
MIDNet w/o $\mathcal{L}_{SSL}$                     &
0.9100                                              &
0.9100                                              &
0.9102                                              &
0.6099                                              &
0.6700                                              &
0.5738                                              &
0.7961                                              &
0.7100                                              &
0.9074                                              \\
MIDNet w/o $\mathcal{L}_{clus}$                       &
0.9421                                              &
0.9425                                              &
0.9466                                              &
0.3406                                              &
0.2350                                              &
0.6586                                              &
0.2264                                              &
0.4900                                              &
0.1472                                              \\
MIDNet w/o $\mathcal{L}_{MI}$                       &
0.9425                                             &
0.9425                                              &
0.9436                                              &
0.5015                                              &
0.4950                                              &
0.5118                                              &
0.7913                                              &
0.7950                                              &
0.7904                                              \\
MIDNet                       &
0.9281                                              &
0.9275                                             &
0.9327                                             &
0.7434                                              &
0.7300                                             &
\textbf{0.7676}                                             &
\textbf{0.8383}                                              &
\textbf{0.8600}                                             &
0.8497                                             \\
MIDNet+$\mathcal{L}_{trip}$                        &
\textbf{0.9649}                                              &
\textbf{0.9650}                                             &
\textbf{0.9658}                                             &
\textbf{0.7768}                                              &
\textbf{0.8500}                                             &
0.7177                                             &
0.6614                                              &
0.6000                                             &
0.8009                                             \\
\bottomrule[1.2pt]
\end{tabular}
}
\end{table*}

\begin{figure*}[t]
    \centering
    \setcounter{subfigure}{0}
    \subfloat[][MIDNet]{
    \begin{tabular}{c}
         \includegraphics[height=3cm, trim=2cm 1.5cm 1.5cm 1.5cm, clip]{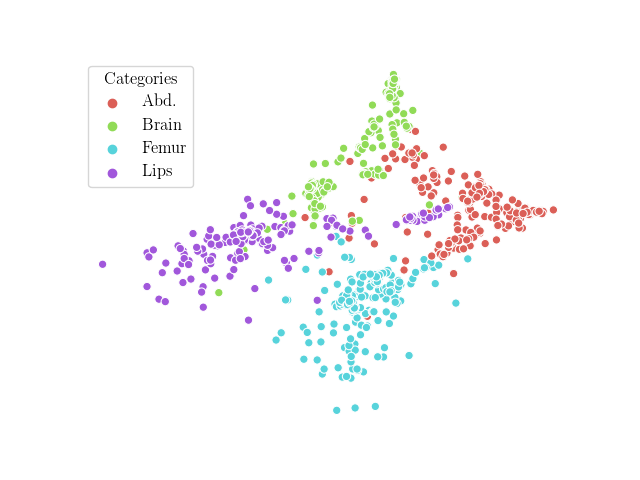}. \\
         \includegraphics[height=3cm, trim=2cm 1.5cm 1.5cm 1.5cm, clip]{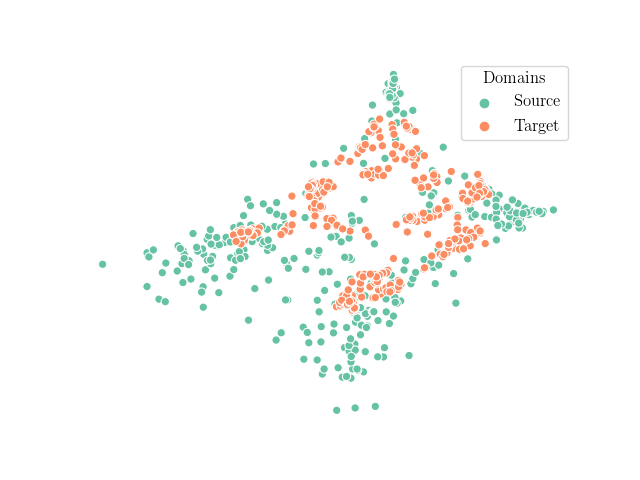}
    \end{tabular}
    }
    \hspace{-0.7cm}
    \setcounter{subfigure}{1}
    \subfloat[MIDNet w/o $\mathcal{L}_{MI}$]{
    \begin{tabular}{c}
         \includegraphics[height=3cm, trim=2cm 1.5cm 1.5cm 1.5cm, clip]{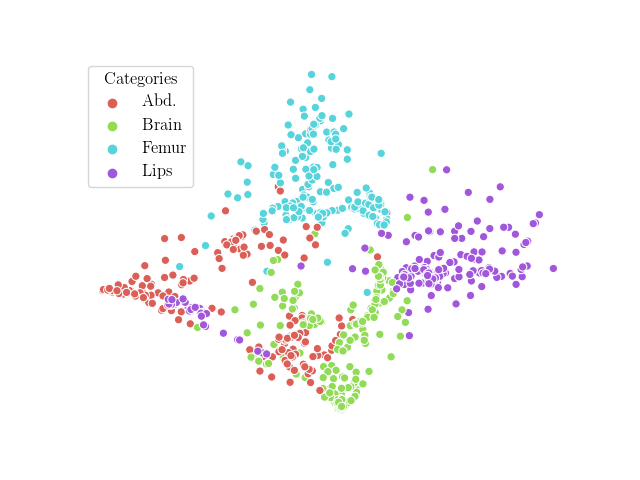}  \\
         \includegraphics[height=3cm, trim=2cm 1.5cm 1.5cm 1.5cm, clip]{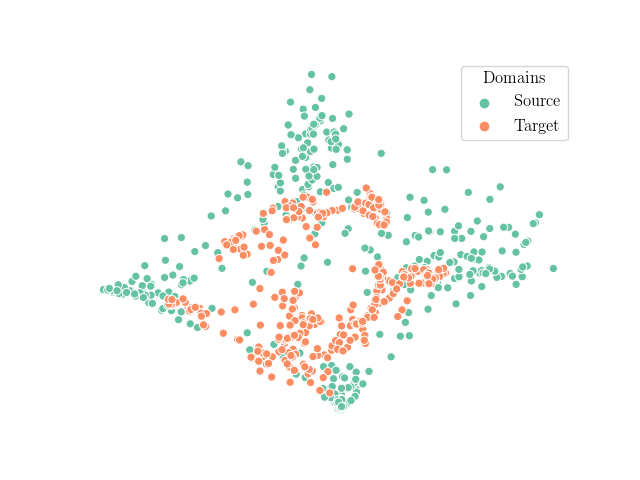} 
    \end{tabular}
    }
    \hspace{-0.7cm}
    \setcounter{subfigure}{2}
    \subfloat[DANN~\cite{Ganin2016}]{
    \begin{tabular}{c}
         \includegraphics[height=3cm, trim=2cm 1.5cm 1.5cm 1.5cm, clip]{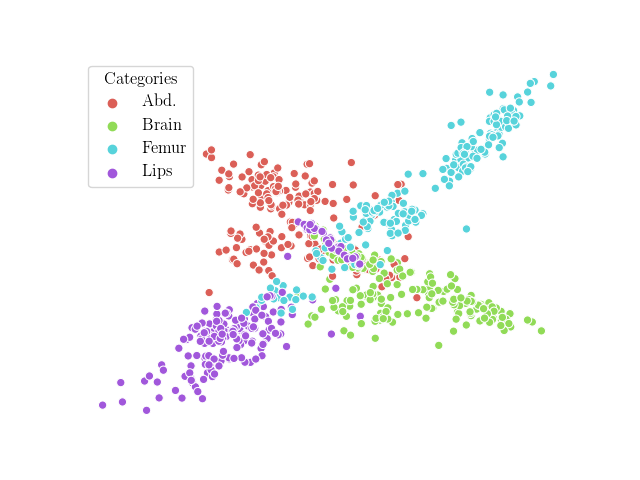}  \\
         \includegraphics[height=3cm, trim=2cm 1.5cm 1.5cm 1.5cm, clip]{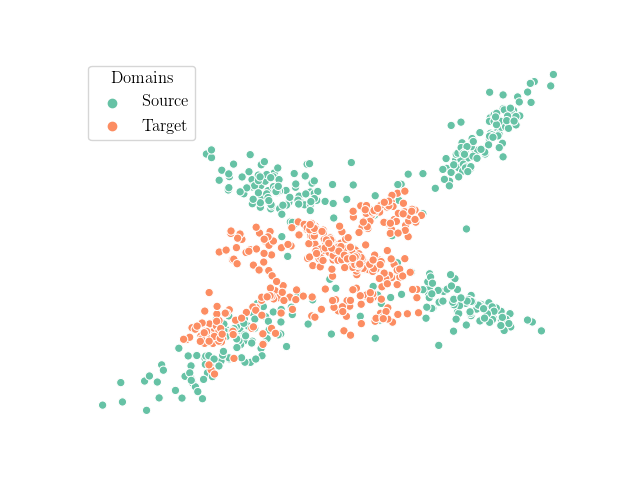} 
    \end{tabular}
    }
    \hspace{-0.7cm}
    \setcounter{subfigure}{3}
    \subfloat[MME~\cite{Saito2019}]{
    \begin{tabular}{c}
         \includegraphics[height=3cm, trim=2cm 1.5cm 1.5cm 1.5cm, clip]{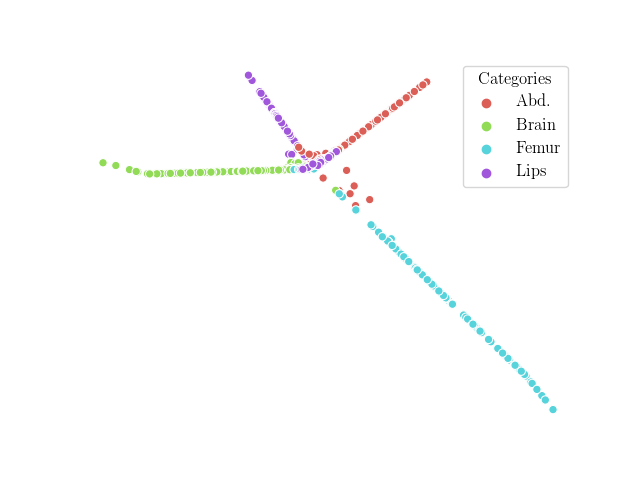}  \\
         \includegraphics[height=3cm, trim=2cm 1.5cm 1.5cm 1.5cm, clip]{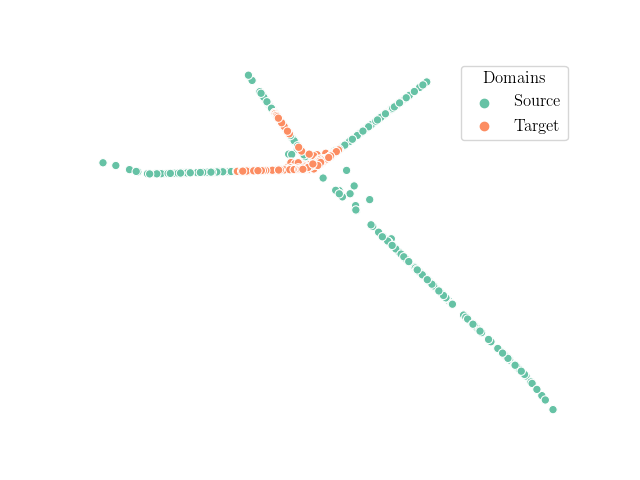} 
    \end{tabular}
    } 
    \caption{Feature visualization with t-SNE. We plot categorical features on both domains. The color in the top row represents categories. The color in the bottom row represents domains. Experiment setting is the same with Table~\ref{US_table_device_A2B}. We observe that the categorical features learned by the proposed method are more domain-invariant and categorical-discriminative than other methods.}
    \label{tsne_12}
\end{figure*}

\subsubsection*{Results}
The classification performance of baselines and the proposed model are shown in Table.~\ref{US_table_device_A2B}. We observe that MIDNet outperforms the state-of-the-art on on the most important test data $T_{Target}^{New}$ for average F1-score and recall. Among all the models in the ablation study, MIDNet+$\mathcal{L}_{trip}$ achieves the best performance on $T_{Source}$ and ${T_{Target}}$, demonstrating that metric learning contributes to the separation of seen categories in both domains. MIDNet outperforms \textit{MIDNet w/o $\mathcal{L}_{SSL}$}, \textit{MIDNet w/o $\mathcal{L}_{clus}$} and \textit{MIDNet w/o $\mathcal{L}_{MI}$} on $T_{Target}^{New}$, illustrating the importance of all proposed components in MIDNet for the classification of unseen categories in the target domain. In addition, Fig.~\ref{Adist_lmi} (b) shows the effectiveness of mutual information disentanglement ($\mathcal{L}_{MI}$) to learn domain-invariant categorical features.

The confusion matrices in Fig.~\ref{CM_MIDNet}(b) show the effectiveness of unlabeled data in the proposed method. The classification accuracy of \textit{Femur} and \textit{Lips} in MIDNet improves when using unlabeled data.

From Fig.~\ref{semi} (b), we observe that classification performance improves with the increase of labeled data in most cases. However, the performance reaches its peak after a certain percentage of labeled data is added. For example, the peak point is $60\%$ in this experiment. 

In addition, we utilize t-SNE plots for feature visualization in Fig.~\ref{tsne_12}. Comparing Fig.~\ref{tsne_12} (a) and Fig.~\ref{tsne_12} (b), we observe that with mutual information disentanglement, (1) samples from the same category are more tightly clustered (see the top row) and (2) the source domain and the target domain are overlap more (see the bottom row). This indicates that mutual information disentanglement is important for learning categorical-focused and domain-invariant features. Fig.~\ref{tsne_12} (a), (c)-(d) show that the proposed method outperforms other state-of-the-art methods for learning category-discriminative and domain-invariant features, especially for unseen categories in the target domain (\emph{e.g.}, (a) vs. (d)).    

\begin{figure}[htb]
 \centering
 \setcounter{subfigure}{0}
 \subfloat[][Source: shadow-free; 
 
 \hspace*{3.3em}Target: shadow-containing]{
 \begin{tabular}{@{\hspace{-1\tabcolsep}}c@{\hspace{0.1\tabcolsep}}c@{\hspace{0.1\tabcolsep}}c@{\hspace{0.1\tabcolsep}}c}
 \raisebox{0\height}{\rotatebox[origin=c]{90}{\makecell{~\scalebox{0.8}{GT}}}}  &
 \raisebox{0.2\height}{\rotatebox[origin=c]{0}{\makecell{~\scalebox{0.8}{4CH}}}}  &
 \raisebox{0.2\height}{\rotatebox[origin=c]{0}{\makecell{~\scalebox{0.8}{Femur}}}}   &
 \raisebox{0.2\height}{\rotatebox[origin=c]{0}{\makecell{~\scalebox{0.8}{Lips}}}}   \\
 \raisebox{1.5\height}{\rotatebox[origin=c]{90}{\makecell{~\scalebox{0.8}{TP}}}}   &
  \stackunder{\includegraphics[height=1.2cm]{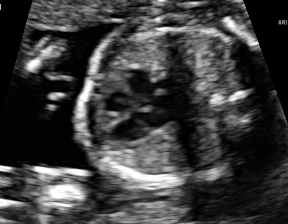}}{~\scalebox{0.8}{4CH}} &
  \stackunder{\includegraphics[height=1.2cm]{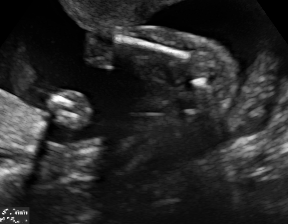}}{~\scalebox{0.8}{Femur}} &
  \stackunder{\includegraphics[height=1.2cm]{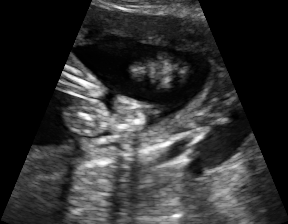}}{~\scalebox{0.8}{Lips}} \\
  \raisebox{1.5\height}{\rotatebox[origin=c]{90}{\makecell{~\scalebox{0.8}{FP}}}}   &
  \stackunder{\includegraphics[height=1.2cm]{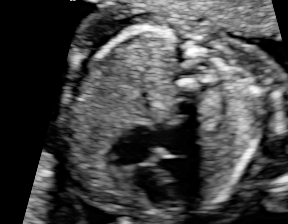}}{~\scalebox{0.8}{Abdominal}} &
  \stackunder{\includegraphics[height=1.2cm]{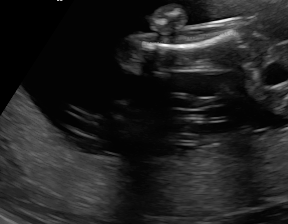}}{~\scalebox{0.8}{Lips}} &
  \stackunder{\includegraphics[height=1.2cm]{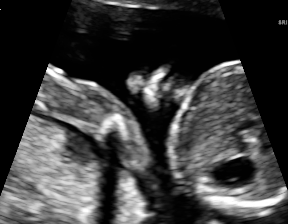}}{~\scalebox{0.8}{LVOT}} 
  \end{tabular}
  }
  \hspace{-0.7cm}
 \setcounter{subfigure}{1}
 \subfloat[][Source: device A; 
 
 \hspace*{1em}Target: device B]{
 \begin{tabular}{c@{\hspace{0.1\tabcolsep}}c}
 \raisebox{0.2\height}{\rotatebox[origin=c]{0}{\makecell{~\scalebox{0.8}{Femur}}}}  &
 \raisebox{0.2\height}{\rotatebox[origin=c]{0}{\makecell{~\scalebox{0.8}{Lips}}}}   \\
  \stackunder{\includegraphics[height=1.2cm]{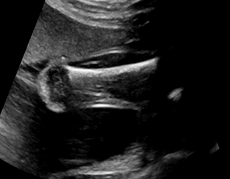}}{~\scalebox{0.8}{Femur}} &
  \stackunder{\includegraphics[height=1.2cm]{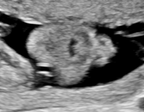}}{~\scalebox{0.8}{Lips}} \\
  \stackunder{\includegraphics[height=1.2cm]{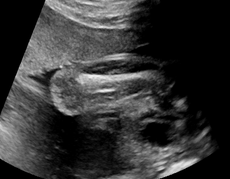}}{~\scalebox{0.8}{Abdominal}} &
  \stackunder{\includegraphics[height=1.2cm]{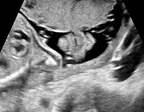}}{~\scalebox{0.8}{Abdominal}} 
  \end{tabular}
  }
  \caption{Examples of classification results on $T_{Target}^{New}$ using MIDNet in Table~\ref{US_table_shadow} and Table~\ref{US_table_device_A2B}. Top row (GT) contains ground truth labels. Middle row (TP) contains true positive results and bottom row (FP) contains false positive results.}
  \label{TPFP_ex}
\end{figure}

We further present correctly classified and mis-classified examples of $T_{Target}^{New}$ using MIDNet in Fig.~\ref{TPFP_ex} (b).

\begin{table}[htb]
\centering
\caption{Average F1-score of the unseen categories in the target domain (\emph{i.e.}, Lips), when an increasing number of common categories between the source domain ($\mathcal{S}$) and the target domain ($\mathcal{T}$) is available during training. Best results in bold. $60\%$ of training data are labeled and the rest are unlabeled.}
\label{sharedclass}
\resizebox{0.5\textwidth}{!}{
\begin{threeparttable}
\begin{tabular}{cc|c|c}
\toprule[1.2pt]
\multirow{2}{*}{Methods}                            & 
\multicolumn{3}{c}{$\mathcal{S}$: Abd., Brain, Femur, Lips}                    \\
\cmidrule{2-4}
~~~~~                 &
$\mathcal{T}$: Abd.                   &
$\mathcal{T}$: Abd., Brain                   &
$\mathcal{T}$: Abd., Brain, Femur              \\
\midrule
MIDNet                       &
0.7816                                             &
0.8047                                              &
\textbf{0.8187}                                             \\
MIDNet+$\mathcal{L}_{trip}$                        &
0.6442                                              &
0.7179                                             &
\textbf{0.7727}                                             \\
\bottomrule[1.2pt]
\end{tabular}
\begin{tablenotes}
\item Abd. is \emph{Abdominal} and hyperparameters are the same as in Sec.~\ref{crossD_exp}.
\end{tablenotes}
\end{threeparttable}
}
\end{table}

\subsubsection*{The influence of common categories}
We further explore the influence of common categories on this cross-device fetal US classification task. We evaluate the classification performance on the unseen category in the target domain (\emph{i.e.}, Lips) with an increased number of common categories in both domains. In this experiment, $60\%$ of training data are labeled data and the rest are unlabeled data. Table.~\ref{sharedclass} shows that the classification performance of the unseen categories in the target domain improves as increasing the number of common categories between the source and the target domain. This might because more common categories can introduce more information about data distributions in the target domain.

\section{Discussion}
\label{discuss}
Current disentanglement methods and domain adaptation methods are able to extract domain-invariant categorical features. However, these methods can hardly transfer knowledge to unseen categories in the target domain because (1) they only consider labeled data and extract domain-invariant features with imitated generalizability (\emph{e.g.}, Two-step-Unfair~\cite{Hadad2018} and Multi-task~\cite{meng2019}), (2) they have model-specific limitations, for example, MME~\cite{Saito2019} optimizes categorical clusters only from available categories of both domains and may negatively affect classification of unseen categories, and (3) they neglect intra-class relationships between both domains. Our method extracts domain-invariant categorical features with a semi-supervised paradigm and further explicitly aligns the these features within each category. 

\begin{figure}
    \centering
    \setcounter{subfigure}{0}
    \subfloat[MIDNet]{
    \begin{tabular}{cc}
         \raisebox{1\height}{\rotatebox[origin=c]{90}{\makecell{~\scalebox{0.7}{\textbf{Source: device A}}}}} &
         \includegraphics[height=2.6cm, trim=2cm 1.5cm 1.5cm 1.5cm, clip]{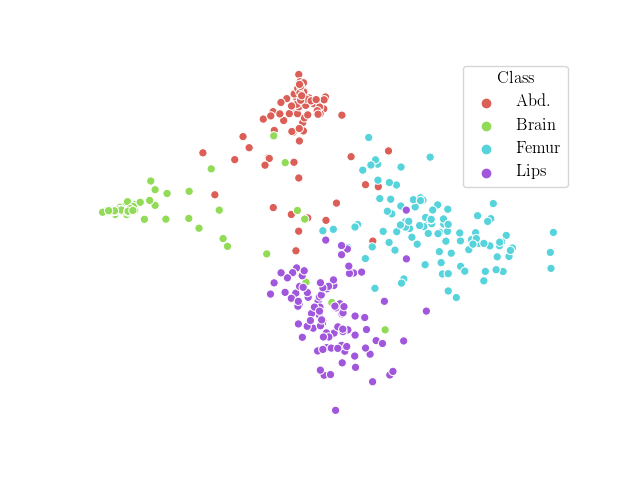}. \\
         \raisebox{1\height}{\rotatebox[origin=c]{90}{\makecell{~\scalebox{0.7}{\textbf{Target: device B}}}}} &
         \includegraphics[height=2.6cm, trim=2cm 1.5cm 1.5cm 1.5cm, clip]{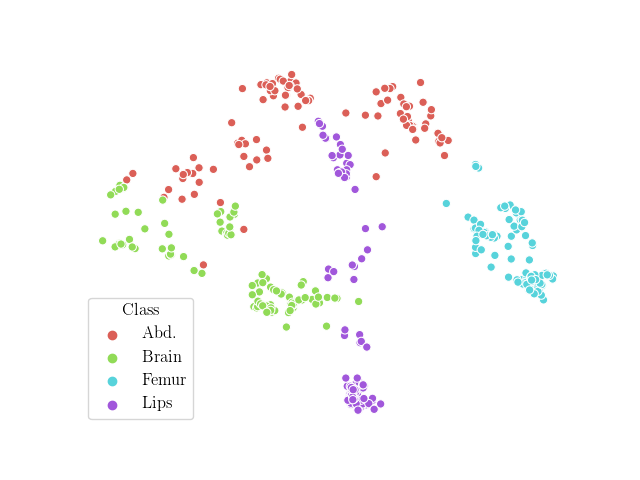}
    \end{tabular}
    }
    \hspace{-0.9cm}
    \setcounter{subfigure}{1}
    \subfloat[MIDNet+$\mathcal{L}_{trip}$]{
    \begin{tabular}{c}
         \includegraphics[height=2.6cm, trim=2cm 1.5cm 1.5cm 1.5cm, clip]{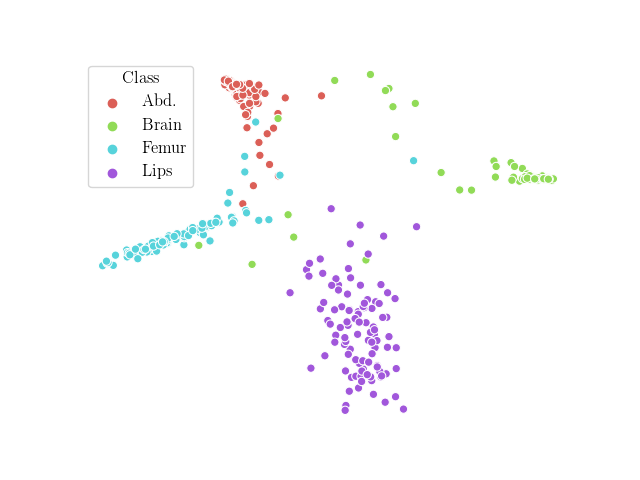}  \\
         \includegraphics[height=2.6cm, trim=2cm 1.5cm 1.5cm 1.5cm, clip]{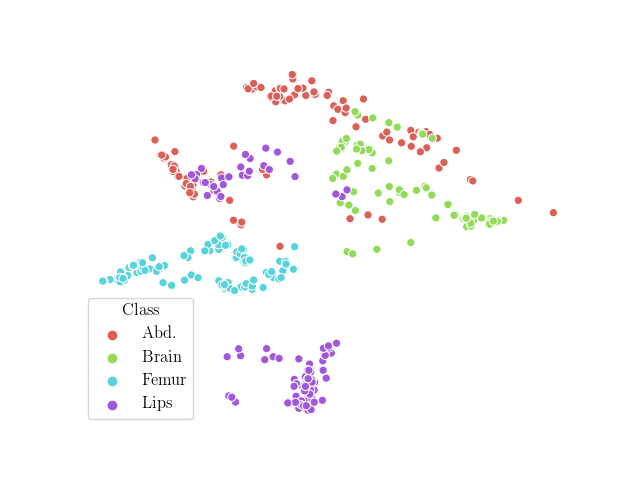} 
    \end{tabular}
    }
    \caption{Comparing categorical feature clusters between MIDNet and MIDNet+$\mathcal{L}_{trip}$. The clusters are visualized by t-SNE plots. The top row is the source domain (device A) and the bottom row is the target domain (device B). We observed that, in MIDNet+$\mathcal{L}_{trip}$, the features of unseen categories are mixed with those of seen categories in the target domain (see the right bottom), although the features of seen categories are more separated.}
    \label{triplet_12}
\end{figure}

The utilization of distance metric learning in feature clustering (Sec.~\ref{FeaCluster}) contributes to further increasing inter-class variance. This results in improved classification performance on $T_{Source}$ and $T_{Target}$ (MIDNet vs. MIDNet+$\mathcal{L}_{trip}$ in Table.~\ref{US_table_shadow} and Table.~\ref{US_table_device_A2B}). However, MIDNet+$\mathcal{L}_{trip}$ models are sometimes outperformed by MIDNet for unseen categories in the target domain $T_{Target}^{New}$. This may be caused by supervised distance metric learning. The distance metric learning encourages latent feature clusters of different seen categories in both domains to move away from each other, shown in the top row of Fig.~\ref{triplet_12}. Such movement of the seen categories may lead to the inter-class mixture between unseen categories and seen categories (\emph{e.g.}, Fig.~\ref{triplet_12} bottom row, lips and femur), and results in increased difficulty for identifying unseen categories in the target domain.

Medical images contain complex entangled image features. For example, shadow artifacts in US imaging are caused by anatomies through blocking the propagation of sound waves or destructive interference.
Traditional DNN-based classifiers jointly learn shadow features and anatomical features without understanding the underlying semantics. 
By observing classification performance on source and target domain after separating entangled image features, our model can be potentially used to interpret the effective factors for target tasks. 
For example, the classification performance of $T_{Source}$ and $T_{Target}$ (Table.~\ref{US_table_shadow} MIDNet and MIDNet+$\mathcal{L}_{trip}$) indicate that shadow features can be more informative for some categories than the actual anatomy in anatomical classification.

The performance of semi-supervised learning usually positively correlates with the percentage of labeled data. In our experimental setting, excessive labeled data may lead to increased class imbalance. This may result in slightly decreased classification performance in Fig.~\ref{semi} (b) when the percentage of labeled training data increases from $60\%$ (i.e., all classes contain the same number of labeled images: 951 images) to $100\%$ (i.e., \textit{Abdominal}: 1448 labeled images vs. \textit{Lips}: 1760 labeled images). A similar observation is made for a digits classification task in Appendix.


The proposed model is effective for learning knowledge from a source domain with a large training set and a target domain with a relatively small training set. This is demonstrated in the cross-device fetal US classification experiment, where the data from the source domain and the target domain contain respectively 2694 and 500 examinations. In the reversed scenario, the decision boundaries that successfully classify categories in a source domain with a small training set might have difficulty to classify unseen categories in a target domain with many samples. This might be the reason for a low classification performance for $T_{Target}^{New}$ when switching the source and the target domains in the cross-device fetal US classification task (Sec.~\ref{crossD_exp}). Nevertheless, our model is meaningful for practical applications since it is reasonable to initially learn from a domain containing data from many patients and subsequently integrate knowledge from secondary datasets with smaller patient populations. 

In experiments where the domain shift is caused by shadow artifacts, shadow-containing and shadow-free images are separated by an expert observer. To evaluate potential inter-observer variability, another two expert observers have been consulted to annotate a subset of 120 images as shadow-containing and shadow-free (10 shadow-containing/shadow-free images are randomly chosen for each category). The results show that $93.3\%$ and $94.2\%$ of the labels from these two experts are identical with the labels from the initial expert observer. A paired sample t-test between the label distributions yields p-values of $0.7973$ and $0.3674$. These high p-values indicate that there is no significant difference between the label distributions, hence the task is relatively straight-forward for skilled sonographers and thus inter-observer variability is negligible. 

Hyper-parameters are highly debated in many related works. Our method has five hyper-parameters which are determined by extensive experiments. We fix $\lambda_1=1$ in Eq.~\ref{Loss} and adjust the other four hyper-parameters as follows. (1) We initialize these parameters according to their importance, \emph{e.g.}, the parameter for the cross-entropy loss is initially relatively high since classification is our main task. (2) We adjust these parameters to ensure that losses decrease as expect on validation data. This adjustment is performed because losses may have different magnitudes, \emph{e.g.}, in Eq.~\ref{Loss}, $\mathcal{L}_{cls}$ is cross-entropy, $\mathcal{L}_{MI}$ is mutual information, $\mathcal{L}_{clus}$ and $\mathcal{L}_{SSL}$ are the Frobenius norm. (3) We run hyper-parameter combinations with grid search according to the results of the previous step, $\lambda_2\in\{10,20,50\}, \lambda_3\in\{10^{-4},5\times10^{-4},10^{-3}\}, \lambda_4\in\{10,50,100\}, \lambda_5\in\{10,50,100\}$. (4) We select the hyper-parameters combination with the best result on validation data.

In this work, we compare the proposed method with the state-of-the-art methods that aim at one-to-one domain adaptation, \emph{i.e.}, require a single source/target domain. We compare two domain adaptation methods which represent two groups of methods. DANN~\cite{Ganin2016} is a typical adversarial-based method, which is widely used and MME~\cite{Saito2019} is a recent semi-supervised method using feature alignment for domain adaptation. Other types of domain adaptation frameworks (shown in Fig.~\ref{DAcomparison}) focus on a different domain adaptation scenario from ours and need multiple source/target domains. For example, domain generalization approaches requires multiple source domains and aim at learning universal knowledge from these multiple source domains. Domain agnostic learning approaches require multiple target domains and focus on transferring knowledge to multiple different target domains. Therefore, to ensure fair comparisons, our work excludes such multi-source/multi-target approaches. As closest related surrogate, we further compare the proposed method with a domain agnostic learning method that uses mutual information (DADA\footnote{https://github.com/VisionLearningGroup/DAL}~\cite{Peng2019} with a single target domain). Table~\ref{vsDADA} shows that our proposed method is better for improving classification performance on unseen categories in the single target domain.

\begin{table}[tb]
\centering
\caption{Classification performance on $T_{Target}^{New}$ with MIDNet and DADA~\cite{Peng2019}. Best results in bold.}
\label{vsDADA}
\resizebox{0.5\textwidth}{!}{
\begin{tabular}{lc|c|c}
\toprule
\multirow{2}{*}{Methods}                            & 
\multicolumn{3}{c}{\textbf{Source: device A; Target: device B}}                    \\
\cmidrule{2-4}
~~~~~                 &
F1-score                   &
Recall                   &
Precision              \\
\midrule
DADA~\cite{Peng2019}                        &
0.6198                                              &
0.6950                                             &
0.6067                                             \\
MIDNet (Ours)                        &
\textbf{0.8383}                                      &
\textbf{0.8600}                                       &
\textbf{0.8497} \\                                     
\bottomrule
\end{tabular}
}
\end{table}

For the classification task with shadow-free/shadow-containing fetal US images, acoustic shadows cause the domain shift. We explore to reduce this effect by $50\%$ dropout in a VGG~\cite{Simonyan15} and Res-VGG~\cite{Simonyan15,He2016}. However, Fig.~\ref{vsDataAugmentation} shows that MIDNet outperforms VGG/Res-VGG+dropout, illustrating that dropout alone has limitation to reduce the domain shift caused by shadows in this task. 

A naive alternative to tackle distribution shift is data augmentation with elastic transformations, including contrast shift, brightness shift and Gaussian blur. Table~\ref{vsDataAugmentation} compares the proposed method with data augmentations. For the classification task with shadow-free/shadow-containing fetal US images, contrast shift and brightness shift are used as data augmentation for VGG~\cite{Simonyan15} and Res-VGG~\cite{Simonyan15,He2016} with dropout. For the cross-device fetal US classification task, contrast shift, brightness shift and Gaussian blur are adopted for data augmentation. Table~\ref{vsDataAugmentation} shows that MIDNet outperforms VGG/Res-VGG+Augment and VGG/Res-VGG+$\text{Augment}^\dag$. This demonstrates that the proposed method is better than naive data augmentation for tackling domain shift, especially for classification of unseen categories in the target domain.

\begin{table}[tb]
\centering
\caption{Comparison between domain-relevant data augmentations (elastic transformation) and MIDNet. Average F1-score is shown in the table. \textit{Augment} includes dropout, contrast shift and brightness shift. \textit{$\text{Augment}^\dag$} includes Gaussian blur, contrast shift and brightness shift. Best results in bold.}
\label{vsDataAugmentation}
\resizebox{0.5\textwidth}{!}{
\begin{threeparttable}
\begin{tabular}{lc|c|c}
\toprule[1.2pt]
\multirow{2}{*}{Methods}                            & 
\multicolumn{3}{c}{\textbf{Source: shadow-free; Target: shadow-containing}}                    \\
\cmidrule{2-4}
~~~~~                 &
$T_{Source}$                   &
$T_{Target}$                   &
$T_{Target}^{New}$              \\
\midrule
VGG+dropout                       &
\textbf{0.5846}                   &
0.6501                                              &
0.6438                                             \\
VGG+Augment                        &
0.5723                                              &
0.6424                                             &
0.7080                                             \\
Res-VGG+dropout            &
0.5003                                             &
0.5988                                              &
0.6878                                             \\
Res-VGG+Augment            &
0.5067                                              &
0.5758                                             &
0.6492                                             \\
MIDNet (Ours)                        &
0.5484                                              &
\textbf{0.6809}                                      &
\textbf{0.7399}                                             \\
\midrule
\midrule
\multirow{2}{*}{Methods}                            & 
\multicolumn{3}{c}{\textbf{Source: device A; Target: device B}}                    \\
\cmidrule{2-4}
~~~~~                 &
$T_{Source}$                   &
$T_{Target}$                   &
$T_{Target}^{New}$              \\
\midrule
VGG+$\text{Augment}^\dag$                        &
0.7515                                              &
0.7016                                             &
0.7232                                             \\
Res-VGG+$\text{Augment}^\dag$                        &
0.9149                                              &
0.5725                                             &
0.5979                                             \\
MIDNet (Ours)                        &
\textbf{0.9281}                                      &
\textbf{0.7434}                                       &
\textbf{0.8383}                                             \\
\bottomrule[1.2pt]
\end{tabular}
\begin{tablenotes}
\item VGG is VGG~\cite{Simonyan15} and Res-VGG is Res-VGG~\cite{Simonyan15,He2016} in Table~\ref{US_table_shadow} and Table~\ref{US_table_device_A2B}.
\end{tablenotes}
\end{threeparttable}
}
\end{table}

\section{Conclusion}
In this paper, we discuss a problem that is rarely evaluated but important in practical scenarios: transferring knowledge from known entangled image features (\emph{e.g.}, categorical features and domain features) to unseen entangled image features (\emph{e.g.}, categories
from a target domain that are not available during training). We propose Mutual Information-based Disentangled Neural Networks (MIDNet) to extract generalizable features, which are essential for such scenarios. Our model is developed with a semi-supervised learning paradigm. Experiments on fetal US images demonstrate the efficiency and practical applicability of our method compared with the state-of-the-art.



\bibliographystyle{abbrv}
\balance

\newpage

\appendix

\subsection{Experiments on handwritten digits data}
\begin{figure}[htb]
 \centering
 \begin{tabular}{@{\hspace{-1\tabcolsep}}c@{\hspace{0.7\tabcolsep}}c@{\hspace{0.7\tabcolsep}}c@{\hspace{0.7\tabcolsep}}c@{\hspace{0.7\tabcolsep}}c@{\hspace{0.7\tabcolsep}}c@{\hspace{0.7\tabcolsep}}c@{\hspace{0.7\tabcolsep}}c@{\hspace{0.7\tabcolsep}}c@{\hspace{0.7\tabcolsep}}c}
  \raisebox{0.5\height}{\rotatebox[origin=c]{90}{\makecell{~\scalebox{0.7}{MNIST}}}} &
  \includegraphics[height=0.8cm]{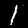} &
  \includegraphics[height=0.8cm]{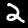} &
  \includegraphics[height=0.8cm]{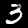} &
  \includegraphics[height=0.8cm]{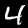} &
  \includegraphics[height=0.8cm]{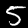} &
  \includegraphics[height=0.8cm]{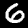} &
  \includegraphics[height=0.8cm]{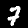} &
  \includegraphics[height=0.8cm]{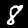} &
  \includegraphics[height=0.8cm]{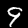} \\
  \raisebox{0.4\height}{\rotatebox[origin=c]{90}{\makecell{~\scalebox{0.7}{MNIST-M}}}} &
  \includegraphics[height=0.8cm]{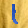} &
  \includegraphics[height=0.8cm]{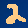} &
  \includegraphics[height=0.8cm]{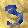} &
  \includegraphics[height=0.8cm]{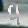} &
  \includegraphics[height=0.8cm]{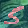} &
  \includegraphics[height=0.8cm]{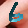} &
  \includegraphics[height=0.8cm]{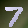} &
  \includegraphics[height=0.8cm]{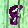} &
  \includegraphics[height=0.8cm]{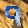}
  \end{tabular}
  \caption{Examples of handwritten digits dataset, containing MNIST~\cite{LeCun1998} and MNIST-M~\cite{Ganin2015}.
  }
  \label{MnistData}
\end{figure}

In this section, we demonstrate the efficiency of our method on an additional handwritten digits  classification task (Fig.~\ref{MnistData}). MNIST is the source domain while MNIST-M is the target domain. Except \textit{Source only}, all the methods are trained on digits 0 to 9 from the source domain and digits 0 to 4 from the target domain. We aim to separate digital features (categorical features) from domain features to obtain generalized digital features, and thus to achieve high digit classification performance on $T_{Target}^{New}$ (digits 5 to 9 from target domain). Here, $T_{Source}$ contains digits 0 to 9 from the source domain and $T_{Target}$ contains digits 0 to 4 from the target domain. In this experiment, we focus on the effectiveness of our method on the unseen categories in the target domain. According to the two applications in the main paper, the feature clustering component mainly improves the classification performance of categories that are available during training. Therefore, we eliminate the feature clustering component in the proposed model and conduct the same evaluation on the digits classification task.  
Hyper-parameters $\lambda_1$ to $\lambda_5$ in Eq.~\ref{Loss} are experimentally chosen as $\lambda_1=1, \lambda_2=10, \lambda_3=10^{-3}, \lambda_4=10^{2}, \lambda_5=10^{3}$.

\textbf{Results:}
The experimental results of baselines and the ablation study are shown in Table.~\ref{mnist_table}. From this table, we observe that the MIDNet model outperforms other baselines on all test data for average F1-score, recall and precision. For example, MIDNet achieves average F1-score of $0.9906$, $0.8204$ and $0.7166$ for $T_{Source}$, $T_{Target}$ and $T_{Target}^{New}$, respectively, while the highest average F1-score of other baselines on the corresponding test data are $0.9802$ (Res-VGG~\cite{Simonyan15,He2016}), $0.7357$ (MME~\cite{Saito2019}) and $0.5794$ (Two-step-fair~\cite{Hadad2018}). Additionally, MIDNet-III performs slightly better than MIDNet on $T_{Target}$ and $T_{Target}^{New}$, demonstrating that feature consistency is important for digit classification. The results of MIDNet-IV and MIDNet-I (similarly, MIDNet-VI vs. MIDNet-II and MIDNet-VIII vs. MIDNet-V) illustrate the effectiveness of SSL based regularization in the proposed MIDNet.

\begin{figure*}[tb]
 \centering
 \subfloat[$\mathbf{T_{Source}}$]{
 \begin{tabular}{c@{\hspace{0.3\tabcolsep}}c}
 \raisebox{0.5\height}{\rotatebox[origin=c]{0}{\makecell{~\scalebox{0.8}{\textbf{With unlabeled data}}}}}  &
 \raisebox{0.5\height}{\rotatebox[origin=c]{0}{\makecell{~\scalebox{0.8}{\textbf{Without unlabeled data}}}}}  \\
  \includegraphics[height=4cm, trim=2cm 0cm 3cm 0cm, clip]{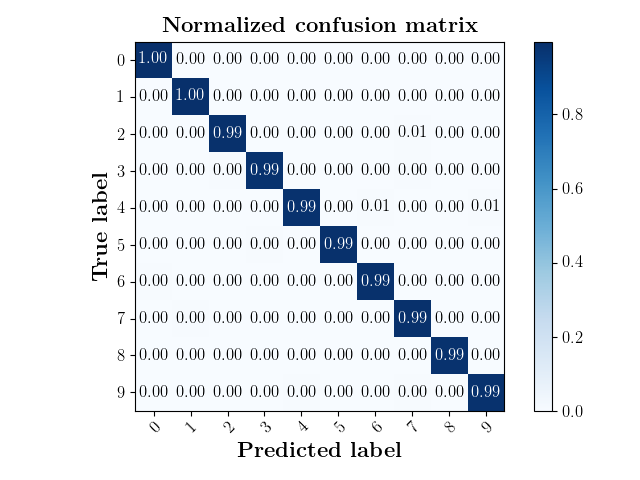} &
  \includegraphics[height=4cm, trim=2cm 0cm 3cm 0cm, clip]{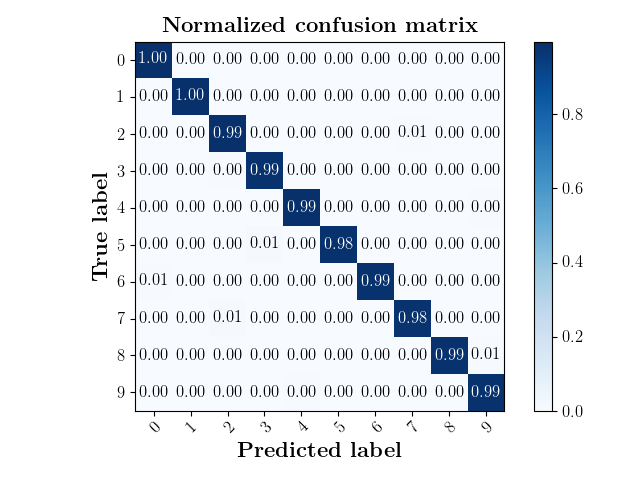}
  \end{tabular}
  }
 \subfloat[$\mathbf{T_{Target}}$ $\&$ $\mathbf{T_{Target}^{New}}$]{
 \begin{tabular}{c@{\hspace{0.3\tabcolsep}}c}
  \raisebox{0.5\height}{\rotatebox[origin=c]{0}{\makecell{~\scalebox{0.8}{\textbf{With unlabeled data}}}}}  &
  \raisebox{0.5\height}{\rotatebox[origin=c]{0}{\makecell{~\scalebox{0.8}{\textbf{Without unlabeled data}}}}}  \\
  \includegraphics[height=4cm, trim=2cm 0cm 3cm 0cm, clip]{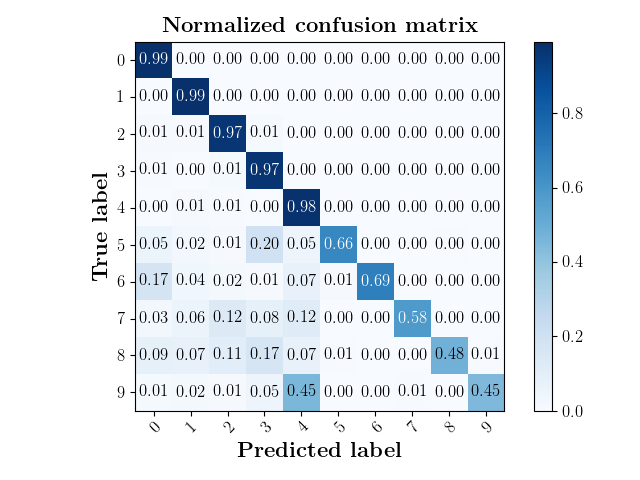} &
  \includegraphics[height=4cm, trim=2cm 0cm 1.4cm 0cm, clip]{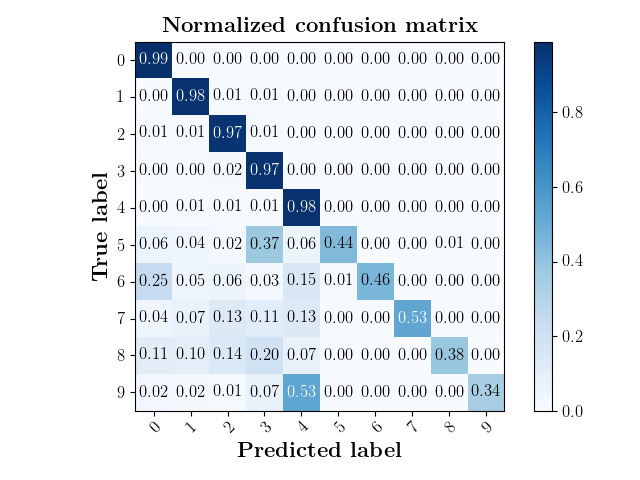}
  \end{tabular}
  }
  \caption{Confusion matrices for digit classification: semi-supervised setting versus fully-supervised setting with MIDNet-VIII model as backbone. For the semi-supervised setting, $30\%$ of training data are labeled data and the rest are unlabeled data. The fully-supervised learning in this experiment only uses the $30\%$ labeled data for training, without using unlabeled data.}
  \label{Mnist_CM}
\end{figure*}

We further compare the performance of MIDNet in a semi-supervised setting and a fully-supervised setting. Here, the semi-supervised setting utilizes the training data containing $30\%$ labeled data and $70\%$ unlabeled data, while the fully supervised setting only uses the $30\%$ labeled data. The confusion matrix in Fig.~\ref{Mnist_CM} shows the effectiveness of unlabeled data in our proposed method, for example, the classification accuracy of $T_{Target}^{New}$ greatly improves when integrating unlabeled data (semi-supervised).

\begin{figure}[htb]
 \centering
 \includegraphics[height=5cm, trim=0cm 0cm 0cm 0cm, clip]{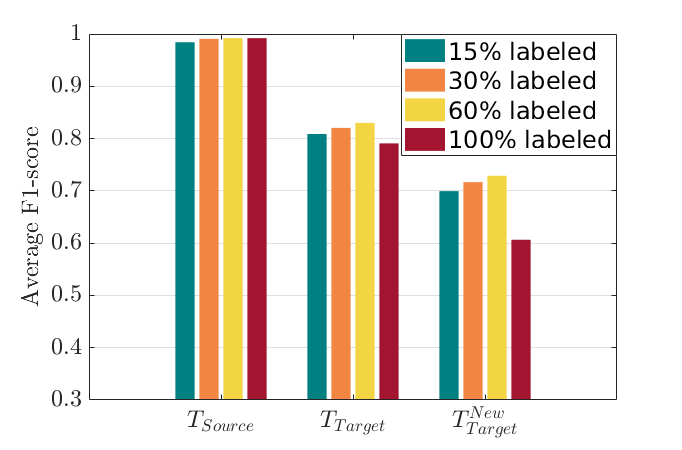}
 \caption{Average F1-score of digit classification with different percentage of labeled data ($15\%, 30\%, 60\%, 100\%$) for semi-supervised learning based on MIDNet model.}
 \label{mnist_semi}
\end{figure}

To explore the importance of labeled data, we evaluate the performance of MIDNet based on using $15\%, 30\%, 60\%$ and $100\%$ labeled data during training. Fig.~\ref{mnist_semi} shows the average F1-score of these experiments on three groups of test data. From this figure, we observe that the classification performance only slightly improves with increasing labeled data. This indicates that MIDNet is capable of achieving expected performance with sparsely labeled data. Additionally, excessive labeled data may lead to increased class imbalance. This may result in decreased classification performance as shown in Fig.~\ref{mnist_semi} when the percentage of labeled training data increased from $60\%$ (i.e., all classes contain the same number of labeled images: 4294 images) to $100\%$ (i.e., \textit{Digit 1}: 10768 labeled images vs. \textit{Digit 5}: 4336 labeled images).

\subsection{Network architectures}

We use residual units for the Encoders $E_1$/$E_2$ and Decoder $D$ of the proposed model.
The implementation of the residual units has been integrated from the publicly available DLTK framework~\footnote{https://dltk.github.io/}. Our implementation is on Tensorflow. The parameter settings (\emph{e.g.}, filters and strides of encoders and decoders, hidden units of the classifier) for these experiments are shown in Table.~\ref{paraSet}.

\begin{table}[htb]
\centering
\caption{The parameter settings of the MIDNet architecture for different experiments. $N$ is the batch size. Digits refers to the handwritten digits classification task that separates digit features from domain features. Fetal US refers to the fetal US standard plane classification tasks that disentangles (1) anatomies from shadow artifacts and (2) anatomies from image acquisition devices.}
\label{paraSet}
\begin{tabular}{ccc}
\toprule
\toprule
~~~~~                            & 
Digits                    & 
Fetal US                   \\
\midrule
Input dimension                   &
($N$, 28, 28, 3, 1)             &
($N$, 224, 288, 1, 1)             \\
Filters ($E_1/E_2/D$)                                   &
(8, 16, 32, 8)                                             &
(8, 16, 32, 64, 8)                                             \\
Strides ($E_1/E_2/D$)                             &
(1, 2, 2, 1)                                     &
(1, 2, 2, 2, 1)                                     \\
Hidden units ($C$)                                  &
(128, 128)                                              &
(128, 128)                                             \\
\bottomrule
\bottomrule
\end{tabular}
\end{table}

\begin{table*}[htb]
\centering
\caption{Comparison of baselines, the state-of-the-art and ablation study (MIDNet-I to MIDNet) for \textbf{digit classification task}. $30\%$ of training data are labeled data and the rest are unlabeled data. Average F1-score, Recall and Precision are measured on three groups of test data. Best results in bold.}
\label{mnist_table}
\resizebox{\textwidth}{!}{
\begin{tabular}{cccc|ccc|ccc}
\toprule[1.2pt]
\multirow{2}{*}{Methods}                            & 
\multicolumn{3}{c|}{$T_{Source}$}                    &
\multicolumn{3}{c|}{$T_{Target}$}                    &
\multicolumn{3}{c}{$T_{Target}^{New}$}              \\
\cmidrule{2-10}
~~~~~                                               &
F1-score                                                & 
Recall                                              & 
Precision                                           &
F1-score                                                & 
Recall                                              & 
Precision                                           &
F1-score                                                & 
Recall                                              & 
Precision                                           \\
\midrule
Source only                                   &
0.9253                                              &
0.9254                                              &
0.9256                                              &
0.5309                                              &
0.5293                                              &
0.5340                                              &
0.5114                                              &
0.5118                                              &
0.5213                                              \\
VGG~\cite{Simonyan15}                                                 &
0.9151                                              &
0.9162                                              &
0.9146                                              &
0.7334                                              &
0.8412                                              &
0.6517                                              &
0.6152                                              &
0.5208                                              &
0.7552                                              \\
Res-VGG~\cite{Simonyan15,He2016}                    &
0.9802                                              &
0.9802                                              &
0.9802                                              &
0.7236                                              &
0.9338                                              &
0.5953                                              &
0.5228                                              &
0.3595                                              &
0.9631                                              \\
Two-step-fair~\cite{Hadad2018}                                       &
0.8704                                              &
0.8707                                              &
0.8704                                             &
0.6908                                              &
0.7806                                              &
0.6203                                              &
0.5794                                              &
0.5002                                              &
0.6911                                              \\
Two-step-Unfair~\cite{Hadad2018}                                       &
0.7465                                              &
0.7492                                              &
0.7591                                             &
0.5839                                              &
0.6407                                              &
0.5428                                              &
0.2983                                              &
0.2598                                              &
0.3894                                              \\
Multi-task~\cite{meng2019}                                          &
0.9318                                              &
0.9315                                              &
0.9332                                             &
0.6203                                              &
0.7824                                              &
0.5171                                              &
0.5053                                              &
0.3695                                              &
0.8368                                              \\
DANN~\cite{Ganin2016}                                          &
0.9678                                              &
0.9679                                              &
0.9681                                              &
0.6818                                              &
0.8901                                              &
0.5579                                              &
0.4506                                              &
0.3023                                              &
0.9091                                              \\
MME~\cite{Saito2019}                                          &
0.9709                                              &
0.9704                                              &
0.9726
&
0.7357                                              &
0.9426                                              &
0.6205                                              &
0.4858                                              &
0.3287                                              &
0.9722                                              \\
\midrule
MIDNet-I                                           &
0.9836                                              &
0.9837                                              &
0.9835                                             &
0.7039                                              &
0.9115                                              &
0.5797                                              &
0.4956                                              &
0.3376                                              &
0.9431                                              \\
MIDNet-II                                           &
0.9841                                              &
0.9842                                              &
0.9842                                             &
0.7059                                              &
0.9160                                              &
0.5809                                              &
0.4916                                              &
0.3322                                              &
0.9501                                              \\
MIDNet-III                                           &
0.9869                                              &
0.9869                                              &
0.9870                                             &
\textbf{0.8333}                                              &
0.9780                                              &
\textbf{0.7298}                                              &
\textbf{0.7511}                                              &
\textbf{0.6137}                                              &
0.9765                                              \\
MIDNet-IV                                           &
0.9858                                              &
0.9860                                              &
0.9859                                             &
0.7439                                              &
0.9569                                              &
0.6169                                              &
0.5207                                              &
0.3566                                              &
0.9771                                              \\
MIDNet-V                                           &
0.9863                                              &
0.9862                                              &
0.9864                                             &
0.8051                                              &
0.9766                                              &
0.6903                                              &
0.6821                                              &
0.5295                                              &
\textbf{0.9807}                                              \\
MIDNet-VI                                           &
0.9868                                              &
0.9869                                              &
0.9868                                             &
0.7532                                              &
0.9602                                              &
0.6253                                              &
0.5541                                              &
0.3900                                              &
0.9689                                              \\
MIDNet-VII                                           &
0.9881                                              &
0.9881                                              &
0.9881                                             &
0.8223                                              &
0.9779                                              &
0.7140                                              &
0.7280                                              &
0.5820                                              &
0.9791                                              \\
MIDNet                                          &
\textbf{0.9906}                                              &
\textbf{0.9905}                                              &
\textbf{0.9906}                                             &
0.8204                                              &
\textbf{0.9803}                                              &
0.7108                                              &
0.7166                                              &
0.5704                                              &
0.9806                                              \\
\bottomrule[1.2pt]
\end{tabular}
}
\end{table*}

\subsection{Clinical Relevance}
In this section, we discuss in detail the connection between our work and clinical scenarios.
The problem of domain shift is probably one of the key reasons hindering deployment of machine learning models in patient care at scale.
Specifically, when a machine learning model is trained in one clinic, it rarely performs well in other clinics where different imaging devices and acquisition protocols are used.
New medical devices, software and hardware alike, have to show consistent performance to be approved by regulators and to become acceptable for wide adoption. 

A naive solution is to train a model on data from potentially thousands of sites but this is infeasible because of regulatory/economical constraints and data sharing limitations. Data often can not be shared because of (1) legality and (2) potential future revenue considerations. 

Another possible solution is to train a new model at each clinical site. However, this solution is infeasible because it requires new clinical trials at high costs and introduces more risks for patients every time a new model is established. 

\begin{figure*}[pht]
    \centering
    \includegraphics[height=4.5cm, trim=1cm 11cm 3cm 1cm, clip]{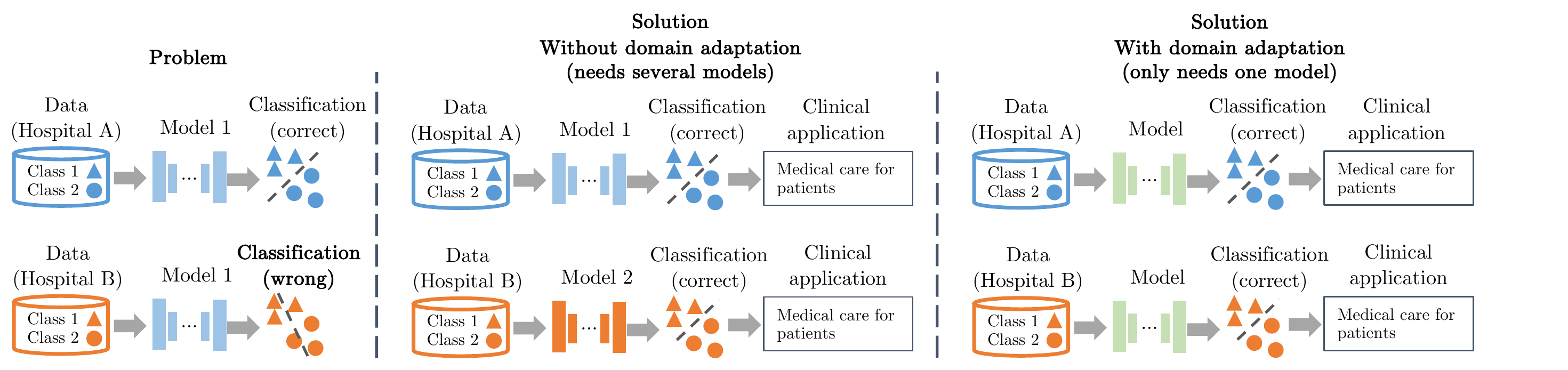}
    \caption{A diagram of the utilization of deep learning models in clinical scenarios with and without domain adaptation.}
    \label{dausage}
\end{figure*}

Therefore, we need domain adaptation methods to reliably transfer models from one clinical site to another and understand their error margins. Fig.~\ref{dausage} shows the utilization of deep learning models in clinical scenarios with and without domain adaptation. 
\begin{figure*}[pht]
    \centering
    \includegraphics[height=9cm, trim=1cm 2.5cm 2cm 1cm, clip]{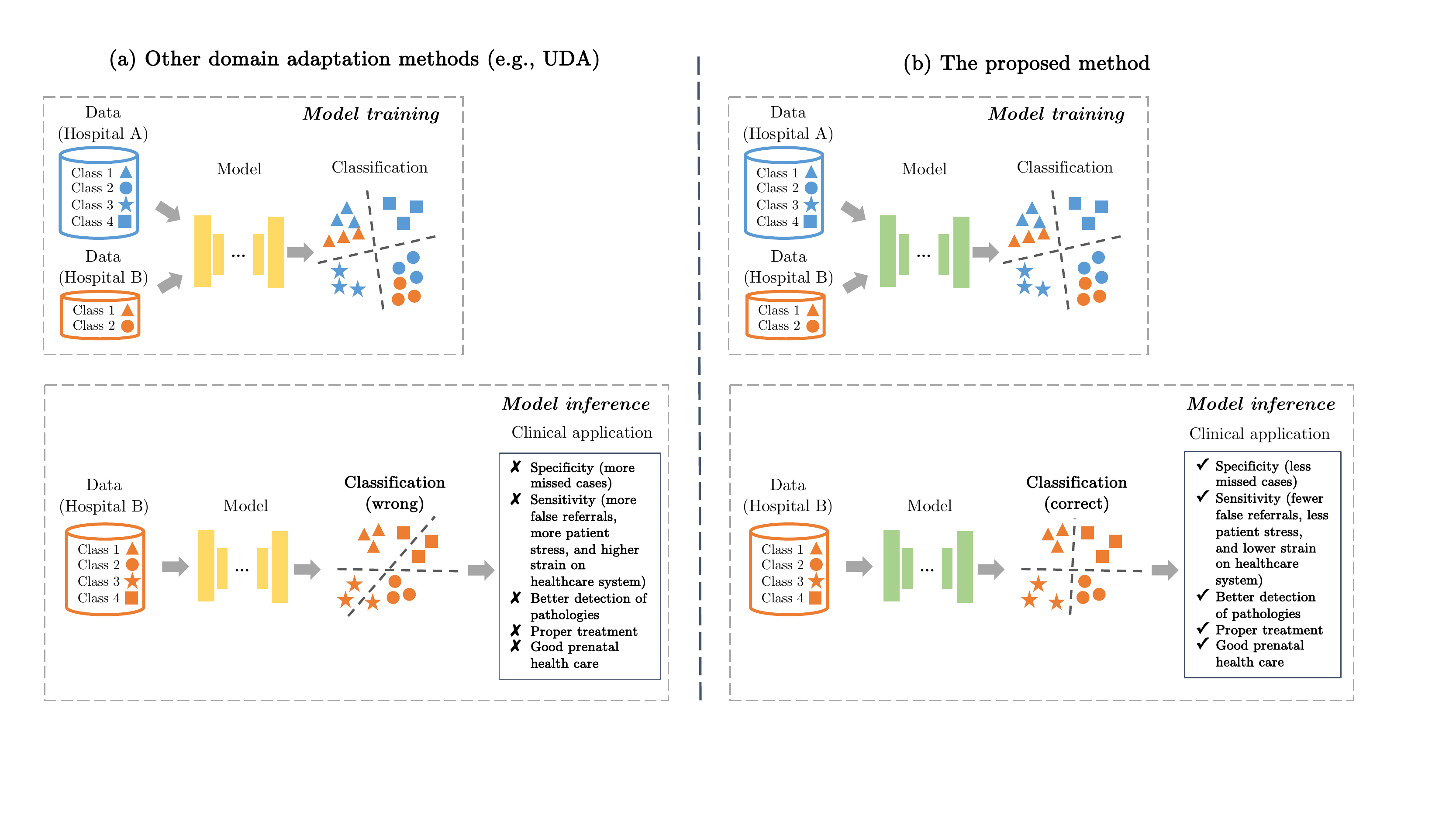}
    \caption{(a) The use of other domain adaptation methods. (b) The proposed method.}
    \label{midnetusage}
\end{figure*}
Generalizing models across datasets with different feature distributions, \emph{i.e.}, domain adaptation, enables a wider and more effective utilization of deep learning models for clinical applications. This helps clinicians from different clinical sites in a wide range of geographic areas to use the same task-specific model for the analysis of their own data (\emph{e.g.}, obtained by different image acquisition devices), and subsequently provide diagnostic decisions and treatment suggestions to their patients.
In the medical image analysis community, many previous studies (\emph{e.g.},~\cite{Kamnitsas2017,Jiang2018,Dou2019,Chartsias2019,Dou2020}) have focused on domain adaptation in different image modalities (\emph{e.g.}, MRI and CT) and organs (\emph{e.g.}, brain and heart) for various clinical applications (\emph{e.g.}, brain lesion and cardiac segmentation). MICCAI (the leading international conference for medical image analysis) has a dedicated session 
and workshops (\emph{e.g.}, DART) about the topic ``domain adaptation''. This demonstrates that domain adaptation addresses a common problem in medical imaging and is important for various clinical applications, and thus is relevant to clinical usage and patient care. 

Specifically in our work, the proposed method is applied to a real-world clinical application, the classification of standardized fetal ultrasound views during prenatal screening. Standardization of anatomical view planes is key to empower the front-line-of-care during screening, making measurements comparable across patients and to accurately predict outcomes. 
We focus on domain adaptation between fetal ultrasound datasets with different feature distributions,  which are caused by imaging artifacts and different acquisition devices. Our study enables learning-based classifiers to be effectively utilized on a wider range of fetal ultrasound images. This helps early detection of pathological development independent from the used imaging setup. The detection of abnormalities can inform downstream treatment decisions and delivery options~\cite{holland2015prenatal}.
Fig.~\ref{midnetusage} illustrates and compares the potential impact of our method and other domain adaptation methods on machine learning for patient care in the context of fetal screening.

\end{document}